\documentclass[letterpaper,twocolumn,10pt]{article}
\usepackage{usenix-2020-09}

\usepackage{tikz}
\usepackage{amsmath}
\usepackage{enumitem}
\usepackage{graphicx}
\usepackage{bm}
\usepackage{multirow}
\usepackage{appendix}
\usepackage{array}
\usepackage{booktabs}
\usepackage{makecell}
\usepackage{diagbox}
\usepackage{soul}
\usepackage{subcaption,graphicx}
\newcommand{\SLJ}[1]{\textcolor{black}{#1}}
\newcommand{\bluetext}[1]{\textcolor{blue}{#1}}
\newcommand{\redtext}[1]{\textcolor{red}{#1}}
\newcommand{\paragraphbe}[1]{\smallskip\noindent{\bf {#1}.}~}
\usepackage{filecontents}

\begin{document}

\date{}

\title{TextDefense: Adversarial Text Detection based on Word Importance Entropy}

\author{
{\rm Lujia Shen}\\
Zhejiang University\\
shen.lujia@zju.edu.cn
\and
{\rm Xuhong Zhang}\\
Zhejiang University\\
zhangxuhong@zju.edu.cn
\and
{\rm Shouling Ji}\\
Zhejiang University\\
sji@zju.edu.cn
\and
{\rm Yuwen Pu}\\
Zhejiang University\\
yw.pu@zju.edu.cn
\and
{\rm Chunpeng Ge}\\
Shandong University\\
gechunpeng2022@126.com
\and
{\rm Xing Yang}\\
National University of Defense Technology\\
yangxing17@nudt.edu.cn
\and
{\rm Yanghe Feng}\\
National University of Defense Technology\\
fengyanghe@nudt.edu.cn
} 


%
%
%
%
%
%
%

\maketitle

\begin{abstract}
Currently, natural language processing (NLP) models are wildly used in various scenarios. 
However, NLP models, like all deep models, are vulnerable to adversarially generated text. 
Numerous works have been working on mitigating the vulnerability from adversarial attacks.
Nevertheless, there is no comprehensive defense in existing works where each work targets a specific attack category or suffers from the limitation of computation overhead, irresistible to adaptive attack, etc. 

In this paper, we exhaustively investigate the adversarial attack algorithms in NLP, and our empirical studies have discovered that the attack algorithms mainly disrupt the importance distribution of words in a text. A well-trained model can distinguish subtle importance distribution differences between clean and adversarial texts.
Based on this intuition, we propose TextDefense, a new adversarial example detection framework that utilizes the target model's capability to defend against adversarial attacks while requiring no prior knowledge.
TextDefense differs from previous approaches, where it utilizes the target model for detection and thus is attack type agnostic. 
Our extensive experiments show that TextDefense can be applied to different architectures, datasets, and attack methods and outperforms existing methods. 
We also discover that the leading factor influencing the performance of TextDefense is the target model's generalizability. 
By analyzing the property of the target model and the property of the adversarial example, we provide our insights into the adversarial attacks in NLP and the principles of our defense method.
\end{abstract}

\section{Introduction}

Deep neural networks (DNNs) have gained tremendous success in many areas, including natural language processing (NLP) \cite{Devlin2019, yang2019xlnet, kale2020text}. 
Such success has facilitated the development of tasks like sentiment analysis, reading comprehension, neural machine translation, etc. 
However, DNNs are known to be vulnerable to malicious perturbations in inputs called adversarial examples. 
For example, the perturbation in the computer vision domain is the modification of pixels. 
Although the inputs of NLP models are texts made of lexical words that are discrete in input space, they also suffer from adversarial attacks. 
Such malicious perturbation in input text can sabotage the text classification models like toxic comment detection, which leads to the proliferation of abusive texts. 

Recent years have witnessed a large number of researches on the design of new attack algorithms for generating adversarial text \cite{li2018textbugger, jin2020bert, ren2019generating, gao2018black, li2020bert, pruthi2019combating}. 
The adversarial attacks in NLP can be classified into three categories, which are character-level, word-level, and sentence-level \cite{wang2019towards}. 
Character-level attacks modify the characters in words to generate adversarial examples. 
However, misspelling checkers can usually defend against this kind of attack. 
Word-level attacks replace words with other words, usually synonyms, to misguide the NLP model without changing the meaning.
They are also the most difficult to defend, which will be elaborated in this paper. 
Sentence-level attacks transform the whole sentence into another with the same semantic meaning, which can be seen as a word-level attack. 
With the development of adversarial attack algorithms, these attacks can be performed under either white-box or black-box settings and with fewer queries and higher semantic similarity, which significantly compromise the security and reliability of NLP model providers.

In the meantime, defense methods have been proposed to protect the NLP models, such as adversarial example detection, adversarial training, model robustness certification, etc. 
Previous detection-based defense algorithms usually target a specific type of attack. 
For instance, DISP \cite{zhou2019learning} and FGWS \cite{mozes2021frequency} are two state-of-the-art (SOTA) detection/restoration methods. 
They first detect adversarial perturbations and then restore them, expecting the restored sentence to be classified correctly. 
Nevertheless, the performance of these methods is restricted by the performance of the detection model.
For instance, the perturbation discriminator in DISP is challenging to identify word-level attacks, and the candidates' selection method in FGWS is only effective for the word-level attack. 
Additionally, recent detection-based methods have gone astray where many complex models or methods are proposed to defend the adversarial example.
However, it is sometimes not that the more complex or larger the model, the better the detection performance. 
Adversarial training can improve the robustness of a model to a certain extent, but with no theoretical guarantee \cite{goodfellow2014explaining},
since the robustness improvement largely relies on the quality and diversity of the generated adversarial texts \cite{sym13030428}. 
As a result, the adversarially trained models are still sensitive to `small' perturbations \cite{10.5555/3367243.3367425, tramer2018ensemble}.
Robustness certification fills the theoretical guarantee gap by calculating the exact robust radius of a sample. 
Such a robust radius can be either used to train a robust model with a theoretical guarantee \cite{fan2021adversarial, Balunovic2020Adversarial} or to evaluate a sample's vulnerability. 
However, the calculation of robustness bound requires heavy computational resources. 
In summary, existing works lack a universal defense method for existing and unknown attacks.
Furthermore, while some methods ensure detection effectiveness, the cost of these methods exceeds that of the NLP model itself (e.g., longer computation time, larger memory resources, etc.).

To address the aforementioned challenges, in this paper, we propose a new approach to detect adversarial examples called TextDefense that is effective on both existing and unknown attacks.
First, to tackle the problem that existing detection methods cannot generalize to new types of attacks, we rethink how adversarial examples are generated in the NLP domain. 
It is known that the adversarial attack algorithm generates adversarial samples by modifying the semantic-important words and most methods rank the importance of words by their influence on the target label.
Essentially, the attack algorithms disrupt the importance distributions of words in a text. 
Specifically, we find that the important words are usually replaced by another rare or nonexistent word, and the model would ignore or misunderstand the word, which decreases the confidence in model predictions.
By continually modifying the semantic-important words, the confidence of the target label gradually decreases until the prediction changes. 
In the iterative process, the importance score of all other words gradually increases, and the importance score of the modified words gradually decreases.
Such a phenomenon is reflected in all attack processes, and we expect to capture some distribution differences led by this phenomenon to detect the adversarial examples. 
Hence, instead of using an auxiliary model to learn the attack pattern and detect perturbations, we utilize the attack model's capability to distinguish adversarial examples. 

Based on our empirical study on the adversarial attacks in NLP, we observe that clean texts perform differently from the adversarial texts on the target model. 
Specifically, we calculate the importance score of each word in the input text and find that clean texts usually have a relatively low scale in their word importance score distribution. 
This means the words in the sentence are closely related to the context, and removing one word will not drastically affect the prediction. 
On the contrary, the adversarial texts usually have higher scales in their word importance score distribution. 
That is to say, the model fails to understand the context by attending to too many irrelevant words, causing the model to misclassify.
Therefore, we use the information entropy to reflect the scale of these distributions. 
To guarantee the efficiency of TextDefense, we sample a portion of words to calculate the importance score.
With several local and adversarial examples, we can determine a threshold that maximizes the F1 score of TextDefense.
Finally, if an input text has an entropy higher than the threshold, we consider it an adversarial example. 

Through intensive empirical evaluations on two model structures and multiple datasets from different tasks, we show that TextDefense is effective in detecting adversarial examples, regardless of attack types.
It also outperforms three SOTA adversarial detection methods that are only effective in specific attack types. 
We also perform different attacking scenarios, including transfer attacks and attacks from shifted data samples, to illustrate the effectiveness of TextDefense. 
We then study the robustness of TextDefense, and the results indicate that our method is robust against adaptive attacks and outperforms existing SOTA adversarial training methods. 
Finally, we discover that TextDefense is highly related to the generalizability of the target model. 

\textbf{Contributions.} In summary, we make the following contributions in this paper.
\begin{enumerate}[leftmargin=*]
\item We propose TextDefense, which to our best knowledge, is the first attack type agnostic adversarial defense against adversarial examples in NLP. 
It utilizes the target model's capability to detect the adversarial examples without training any auxiliary model for detection and restoration. 
\item We evaluate the effectiveness of TextDefense on different datasets from two tasks (sentiment analysis and toxic comment detection) under multiple attack methods (TextBugger, TextFooler, PWWS, etc.). 
The evaluation results indicate that TextDefense outperforms existing methods and is effective in different settings.
Meanwhile, our experiment on online NLP APIs shows that TextDefense can enhance the security of MLaaS.
\item We verify the robustness of TextDefense on adaptive attacks and compare it with state-of-the-art adversarial training methods. 
The results show that TextDefense can depress the attack success rate and increase the number of perturbed words required for a successful attack.
\end{enumerate}

\section{Related Work}
\subsection{Adversarial Attack in NLP}
Recent works have shown that DNNs are vulnerable to adversarial attack \cite{goodfellow2014explaining} where imperceptible perturbations are generated maliciously to cause the DNN model to misbehave (e.g., misclassification). In the computer vision (CV) domain, perturbations are the modified pixels generated using methods like projected gradient descent \cite{madry2018towards}. 
Unlike CV, adversarial attacks in NLP modify words in the input text to achieve the purpose of the attack \cite{li2018textbugger, jin2020bert, ren2019generating, alzantot2018generating, li2020bert, garg2020bae, gao2018black}. 

In the NLP domain, the adversarial attack can be classified into three categories which are character-level attack, word-level attack, and sentence-level attack \cite{wang2019towards}. 
Gao et al. \cite{gao2018black} proposed a character-level attack, called DeepWordBug, to generate adversarial examples in the black-box scenario. 
They first quantified the importance of words and then added imperceptible perturbations to the selected words through swapping, flipping, insertion, and deletion. 
In a character-level attack, only characters in the word will be perturbed.

Unlike character-level attacks, word-level attacks replace the whole word to achieve the attack goal. 
Inspired by the fast gradient sign method (FGSM), Papernot et al. \cite{papernot2016crafting} generated adversarial examples by calculating the model's gradient in a white-box scenario. 
However, the selected substitution words using FGSM may have grammatical errors. 
Recent works have focused on the word-level attack by synonym substitution, which is more stealthy.
Jin et al. \cite{jin2020bert} presented TextFooler, a black-box attack to fool BERT \cite{Devlin2019} on text classification. 
They first identified the important words for the target model and then replaced them with synonyms. 
Similarly, Ren et al. \cite{ren2019generating} designed PWWS to generate adversarial examples by synonym substitution. 
PWWS and TextFooler only differ in how synonyms are chosen.
Li et al. \cite{li2018textbugger} proposed TextBugger, a multi-level attack consisting of both character-level and word-level attacks.

The sentence-level attack usually inserts sentences at the beginning, middle, or end of the text with the semantics and grammar unchanged. This kind of attack usually appears in other NLP tasks like neural machine translation, reading comprehension, and question answering \cite{wang2019towards}.

\subsection{Defense Methods} \label{sec:ba}
We classify the defenses into three categories: adversarial example detection/restoration, adversarial training, and robustness certification. 

\paragraphbe{Detection/Restoration Method} We group the detection and restoration methods together because detection usually appears together with restoration. 
Zhou et al. \cite{zhou2019learning} proposed a framework called DISP to first detect possible perturbations in the input and then restore the detected perturbations. 
DISP can detect character-level perturbations because misspelled words are out-of-vocabulary tokens that can be located easily by its discriminator. 
However, word-level attacks like synonym substitution are challenging to detect and restore for them. 
Mozes et al. \cite{mozes2021frequency} proposed FGWS first to identify the rare words in the input and then replace them with their synonym that has higher occurrence. 
Nevertheless, FGWS performs better in word-level attacks where common words are usually replaced by their rare synonyms.
Nevertheless, character-level attacked words are not real words. 
Thus, FGWS cannot find their corresponding synonyms and fails to detect character-level attacks.
Li et al. \cite{li2020textshield} proposed TextShield to defend adversarial examples. 
The core part of TextShield is a neural machine translation (NMT) model that converts adversarial input into its original format. 
Nonetheless, prior knowledge of the possible attack types is required to train a good NMT model. 
If a new type of attack is proposed, TextShield will not be able to restore samples from this type of attack.
In summary, current detection and restoration defenses are limited to specific attack types. 
We will compare the above-mentioned three defense methods in our experiments.

\paragraphbe{Adversarial Training}
Adversarial training is widely applied to defend against adversarial examples in both CV and NLP. 
Researchers mix adversarial examples with the original ones for retraining to improve the models’ tolerance to adversarial examples. 
Previous adversarial training heavily relies on human knowledge or suffers from low diversity of attacks, which limits the robustness to attacks with diverse words and expressions. Therefore, Xu et al. \cite{xu2019lexicalat} proposed LexicalAT, which implements reinforcement learning with the adversarial attack to generate more robust adversarial examples. 
Zhou et al. \cite{zhou2021defense} proposed Dirichlet Neighborhood Ensemble (DNE) to defend against synonym substitution attacks. 
They first used randomized smoothing to estimate the prediction and then trained the model with two-hop neighbors. 
Yoo et al. \cite{yoo-qi-2021-towards-improving} proposed a computationally cheaper adversarial training framework A2T, which generates adversarial examples based on synonym substitution. 
However, the adversarially trained models are still vulnerable to adversarial attacks. 
Moreover, adversarial training can hardly capture all kinds of word-level perturbations.

\paragraphbe{Robustness Certification}
Recent researchers proposed robustness certification to search the boundary to guarantee that the sample is robust to attacks within the boundary. 
Ko et al. \cite{ko2019popqorn} proposed POPQORN to find the lower bounds for robustness quantification. 
They leveraged interval arithmetic to approximate the non-linearity of RNNs. 
Nevertheless, it is inefficient and poses a computational challenge for a larger RNN model. 
Du et al. \cite{du2021cert} proposed to use abstract interpretation to estimate the perturbation space through the model, which reduces the computation overhead and maintains the precise estimation. 
Nevertheless, it still requires much more computation resources due to the bisection method.
Another robustness certification method is randomized smoothing \cite{cohen2019certified} which takes samples around the input sample with normal distribution to get a robustness radius of the input sample. 
However, the normal distribution around NLP input is not practical due to the discreteness of the input space.

\section{Method}

\subsection{Problem Definition and Threat Model}
Given an input text $x$ that contains $n$ words (i.e., $x ={w_1,w_2,\cdots ,w_n}$), and an NLP model $F: X\rightarrow Y$ which maps from the input space $X$ to the label space $Y$, an attacker who has access to the model $F$, regardless of black-box or white-box, aims to generate an adversarial text $x'$ from $x$ whose ground truth label is $y\in Y$, such that $F(x') = t$ where $t \ne y$. 

In this paper, we aim to defend against such attacks by detecting whether the input is adversarial or not. We make use of the capability of the target model itself to determine the importance score distribution of the text. After quantifying the distribution, we can discriminate the adversarial examples with a threshold precomputed using a few local samples. 

\begin{figure}[h] 
    \centering
    \includegraphics[scale=0.38]{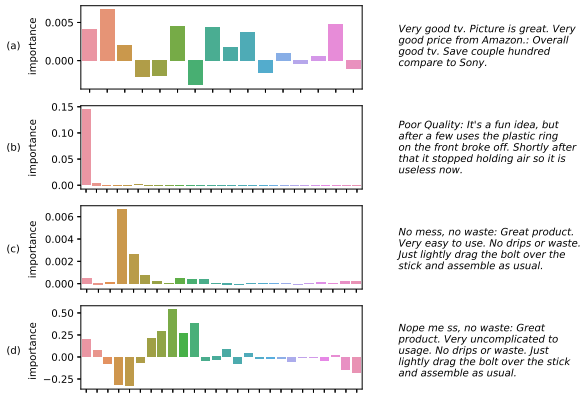}
    \caption {Illustration of three different score distributions.}
    \label{fig:illustration}
    \vspace{-7mm}
\end{figure}

\begin{figure}[t] 
    \centering
    \includegraphics[scale=0.65]{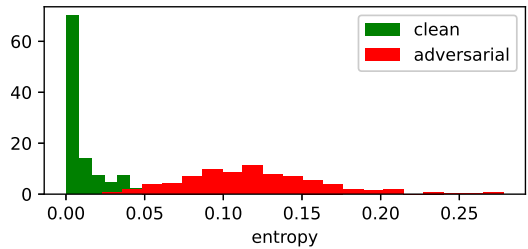}
    \vspace{-4mm}
    \caption{The histogram of the entropy values of importance scores for clean texts and adversarial texts.}
    \label{fig:densityplot}
    \vspace{-4mm}
\end{figure}

\subsection{Design Intuition}
Before introducing our method, we would like to share some insights into adversarial example generation, further inspiring us to propose TextDefense.

As we investigate the recent black-box and white-box attack algorithms, we find that the first step for these algorithms is to seek out the vulnerable words according to their importance score \cite{li2018textbugger,jin2020bert,ren2019generating, garg2020bae, li2020bert, gao2018black}. 
Then, they iteratively modify the words one by one based on the ranking of their importance scores using character-level or word-level attacks. 
In the iterative process, the confidence of the model prediction gradually decreases until the prediction changes. 
Specifically, the character-level attacks would replace the task-related words with non-existing words, which would be tokenized into the unknown token. 
The word-level attacks would replace the words with complex words or neutral words. 
For the attack from the unknown token, the model loses its attention to task-related words, causing the effect of irrelevant words to be amplified and thus leading to the change in prediction.  
For the attack from complex words, since the model does not learn the knowledge of the complex words, the model cannot understand the meaning of the text and then misclassify the text.
At the same time, similar to the attack from the unknown token, the model would pay more attention to the irrelevant words.
For the attack from neutral words, such an attack would result in a semantics change of the text, which is no longer considered a valid adversarial attack.
Hence, we make use of the magnification of the model's attention toward irrelevant information to identify adversarial examples.

We implement the same black-box word importance score calculation method as in \cite{li2018textbugger, jin2020bert,garg2020bae} to examine the importance score for all words in a sentence. 
We discover that the score distributions of clean texts have relatively low scales, while the score distributions of adversarial examples have higher scales.
In detail, firstly, we observe that in some clean texts, the importance scores of all words are close to 0, as shown in Fig. \ref{fig:illustration}a. 
That is, removing any word will not change the output logits. 
These texts usually contain multiple task-related words, and the model takes advantage of the task-related information from multiple sources. 
Therefore, these texts are not sensitive to single word deletion and are more robust against adversarial attacks.
From the point of view of human comprehension, the context in the text is closely related and coherent, which allows us to determine the meaning of the text from multiple perspectives.
Secondly, in other clean texts, there exist some words with high importance scores, while the remaining words have scores close to zero, as shown in Fig. \ref{fig:illustration}b. 
These words with high importance scores are all task-related words that the model has learned.
This indicates that the model understands both the context and the task-related words, leading to the correct prediction.
From the human comprehension perspective, the key words and context are coherent, from which we can infer the meaning of the text.
Thirdly, in contrast, we discover that many words in an adversarial text have high importance scores, including both task-relevant words and task-irrelevant words, as shown in Fig. \ref{fig:illustration}d whose corresponding clean distribution is shown in Fig. \ref{fig:illustration}c.
From the robustness perspective, adversarial texts are susceptible to these irrelevant words. 
This indicates that the model's prediction and irrelevant words are highly coupled.
From the human comprehension perspective, we need to infer the meaning of an adversarial text from many irrelevant words, making the meaning less coherent.

To illustrate the difference between these distributions, we calculate the absolute entropy value of the importance scores for clean texts and their corresponding adversarial examples and show their distributions in Fig. \ref{fig:densityplot}.  
We can see a huge difference between the distribution of the entropy values for the clean and adversarial texts. 
The entropy values of the importance scores of the clean texts are close to 0, whereas the values of adversarial texts are large and diverse.
This is consistent with the intuition that a more coherent text has lower entropy.

\begin{figure}[t] 
    \centering
    \includegraphics[scale=0.32]{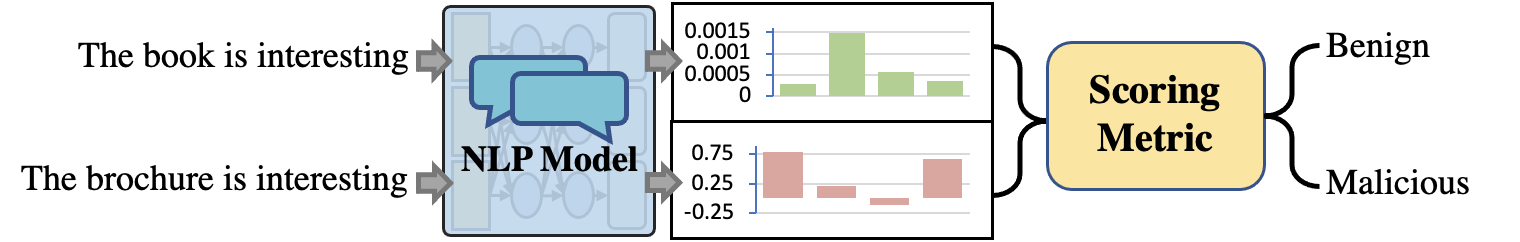}
    \caption{The overview of TextDefense.}
    \label{fig:overview}
    \vspace{-4mm}
\end{figure}

Based on the discrepancy mentioned above, we can formulate our detection method, which we will elaborate below. The overview of our method is shown in Fig. \ref{fig:overview}.

\subsection{Defense Method}\label{sec:method}
\paragraphbe{Word Importance Rank}\label{sec:wir}
Primarily, we have an NLP text classifier $F$ that needs to be protected, an input text $x$, and the target label $y$. 
The classifier outputs the confidence for $k$ labels: 
$$F(x)=[c_1,\dots, c_k]$$
where $\sum_{i}{c_i}=1$. We denote the confidence for the target label $y$ as $F_y(x)$ which equals to $c_y$. Given a text $x=[w_1, w_2, \dots, w_n]$, we use word importance rank (WIR) algorithm with deletion to calculate the importance scores for all words $w_i$ in a sentence $x$.
The importance score of word $w_i$ towards the target label $y$ under the WIR algorithm with deletion is calculated by 
$$s_i=F_y(w_1,w_2,\dots,w_n)-F_y(w_1, \dots, w_i-1,w_i+1,\dots, w_n) .$$

The reason for using the WIR algorithm is that current transformer-based models (e.g., BERT, XLNet, GPT, etc.) utilize the tokenization method (i.e., WordPiece \cite{schuster2012japanese}) to construct the vocabulary of the model instead of using the whole English vocabulary like the RNN-based models (e.g., vanilla RNN, GRU, LSTM, etc.).
This tokenization method would convert complex words into short tokens, which can greatly reduce the number of words in the vocabulary. 
However, the previous gradient-based method cannot fully represent the importance of a word, as some words will be tokenized into multiple tokens.
Thus, we use the WIR with deletion algorithm instead of the gradient-based method to calculate the importance score of a word. Nevertheless, we further study using the gradient method in Appendix \ref{sec:gradient}. 

\paragraphbe{Scoring Metrics}
Now, we have the score distribution for the whole sentence $x$ which is $S=[s_1,s_2,\dots,s_n]$ where $s_i\in[-1,1]$. 
The importance score $s_i$ of the word $w_i$ towards the target label $y$ can be considered as the word's information with respect to the target label compared to the remaining part of the text. 
Based on the above findings, we expect to construct a metric that can distinguish the distributions with low scales and those with high scales. 
We know that a higher positive importance score indicates that the word contributes more to the predicted label; that is, the word contains more information concerning the target label.
On the contrary, if the word has a high negative importance score, the word contains more information about other labels than the target label. 

Therefore, we can use entropy to quantify the information of all words to represent the total information the text contains. 
Due to the existence of negative importance scores, we calculate the entropy for the absolute importance scores. 
Hence, the entropy $H$ of an input text is calculated as
$-\sum_{i=1}^{n}{|s_i|\log|s_i|}.$

Then, a higher entropy value indicates that the text contains too much redundant information that a normal text would not have. 
A lower entropy value indicates that the information in the text is concentrated.
Finally, we select several local samples and generate their adversarial examples using any adversarial attack method, and then determine the threshold with the highest F1 score to discriminate adversarial texts from clean texts. 
In Appendix \ref{appendix:threshold} and \ref{appendix:threshold2}, we have shown that attack methods and numbers of local adversarial examples will not affect the choice of the threshold. 
Thus, any local adversarial attack method can be used to determine the threshold.
Since calculating the importance score for all words requires heavy computational resource, we study the influence of sampling technique on TextDefense's performance in Appendix \ref{sample} and conclude that randomly sampling fewer words from the text can get an entropy value close to the original value.
We also employ other metrics to quantify the scale of the importance scores and evaluate the performance of TextDefense in Appendix \ref{sec:scoring}.

\section{Experimental Settings}

\subsection{Models}\label{models}
In our experiments, we mainly use the BERT model \cite{Devlin2019} for evaluation. 
We also evaluate the LSTM model \cite{LSTM} which has been less used in NLP tasks recently. 
For the BERT model, we use the uncased BERT base model (12-layer, 768-hidden, 12-head, 110M parameters), which is pre-trained on lower-cased English text. 
For text classification, the model will predict the output representation of {\fontfamily{qcr}\selectfont[CLS]} using a multilayer perceptron into label space. 
For the LSTM model, we use the 1-layer bidirectional LSTM with a 400k vocabulary and 150-dimensional hidden state.
In the classification task, the rightmost hidden representation is used to predict the label using a multilayer perceptron.
The parameters of the LSTM model are randomly initialized and trained on specific datasets directly. 

\subsection{Datasets}
For binary classification, we evaluate TextDefense on three sentiment analysis datasets and two toxic comment detection datasets. 
The three sentiment analysis datasets are Amazon, Yelp, and IMDb. 
The Amazon dataset contains product reviews. The Yelp dataset contains crowdsource reviews about various business sources. The IMDb dataset contains movie reviews.
The two toxic comment detection datasets are Twitter and Jigsaw. 
The Twitter dataset~\cite{founta2018large} contains a large number of tweets crawled from Twitter and their abusive behavior annotated humans. 
The Jigsaw dataset contains comments from Wikipedia’s talk page, and the toxicity is rated by crowd-workers.

For multi-class classification, we use Emotion and AG News datasets. 
Emotion is a dataset of English Twitter messages with six basic emotions: anger, fear, joy, love, sadness, and surprise. 
AG News dataset is constructed by assembling titles and descriptions of articles from the four classes: World, Sports, Business, Sci/Tech.
 
For each dataset, we randomly sample 500 texts and generate their adversarial examples. The generated adversarial examples and the corresponding original clean texts are used to evaluate the performance of TextDefense.

\subsection{Metrics}
\paragraphbe{Attack Success Rate (ASR)} This metric is widely used for evaluating the performance of attack algorithms. It is defined by the proportion of successful attack samples in all samples. A higher ASR indicates the attack algorithm is stronger or the defense method is weaker. 

\paragraphbe{Area Under Curve (AUC)} To remove the dependence on a pre-defined threshold, we report the commonly used AUC as the metric of the defense performance. 

\paragraphbe{Perturbed Words (PW)} Since text is discrete data, where a pre-defined embedding vector represents a word, we cannot use metrics used in Euclidean space to measure the perturbations. 
Hence, we use the percentage of required perturbed words to quantify the noise scale in adversarial texts.

\paragraphbe{Semantic Similarity (SIM)} \SLJ{Semantic Similarity is a metric that measures how similar two sentences are. 
We calculate the hidden representations of each original text and its adversarial text using a pre-trained sentence encoder model\footnote{https://huggingface.co/sentence-transformers/all-MiniLM-L6-v2} and apply cosine similarity to get the semantic similarity between the two texts. 
According to our empirical studies, we have found that similarity over 0.9 indicates that the meaning of the generated text barely changed.
Then, a similarity between 0.8 to 0.9 indicates that the generated text is not very consistent with the original text. 
Finally, a similarity below 0.8 indicates that the generated text has changed the original text's meaning.
}

\paragraphbe{Linguistic Acceptability (LA)} The adversarially generated texts are usually grammatically inconsistent or context incoherent. 
\SLJ{We use a classification model\footnote{https://huggingface.co/textattack/bert-base-uncased-CoLA} trained on Corpus of Linguistic Acceptability (CoLA) \cite{warstadt2019neural} to predict whether the text is acceptable or not. 
We use the percentage of acceptable texts to quantify the grammatical quality of the adversarial texts.}

\subsection{Attack Method}
Recent studies have shown that character-level attacks can be defended by grammar detection or spelling checker, whereas the adversarial examples created by word-level attacks are more imperceptible for humans and more difficult for DNNs to defend~\cite{du2021combating, wang2020defense}. 
Therefore, in the experiments, we focus more on word-level attacks where we include three attacks containing word-level attacks. 
We also reflect character-level attacks in our experiments using the attack of TextBugger.

\paragraphbe{TextBugger} Li et al. \cite{li2018textbugger} proposed a multi-level attack framework TextBugger. The words are first ranked by the gradient in white-box settings and WIR with a deletion in black-box settings. 
Then, the words are modified using character-level and word-level attacks successively. 
The character-level attack includes insertion, deletion, swapping, and character substitution, and the word-level attack is word substitution with the nearest neighbors in context-aware word space.

\paragraphbe{TextFooler} Jin et al. \cite{jin2020bert} proposed a word-level attack framework TextFooler. Like TextBugger, they first identify the important words for the target model and then replace them with synonyms until the prediction is altered. 
The synonyms are generated based on counter-fitting word vectors.

\paragraphbe{PWWS} Ren et al. \cite{ren2019generating} designed a word-level attack framework PWWS. Like TextFooler, which replaces words with synonyms based on counter-fitting word vectors, PWWS uses WordNet to choose candidate synonyms. 

The above-mentioned three attacks are all untargeted attacks with minimal perturbation, as this attack goal can make the attacked examples more imperceptible.

\subsection{Baseline Defense Method} \label{sec:baseline}
We implement and compare TextDefense with the other three methods, which are DISP \cite{zhou2019learning}, FGWS \cite{mozes2021frequency}, and TextShield \cite{li2020textshield}. They are three SOTA adversarial examples detection/restoration methods.

\paragraphbe{DISP}
In \cite{zhou2019learning}, they first identified adversarial perturbations using a perturbation discriminator, a BERT model trained on perturbed texts. 
Then, for each identified perturbation, an embedding estimator is learned to restore the word from perturbation based on the context. 

\paragraphbe{FGWS}
In \cite{mozes2021frequency}, they discovered that word substitutions are identifiable through frequency differences between the replaced words and their corresponding substitutions. 
Therefore, they replaced rare words in the text with synonyms with higher occurrence frequencies.
Finally, if the change of prediction score has exceeded a threshold, it is considered an adversarial example.

\paragraphbe{TextShield}
In \cite{li2020textshield}, they used a neural machine translation (NMT) model to translate the adversarial text into its original text. 
Similar to FGWS, the input is considered an adversarial example if the change of prediction score has exceeded a threshold.

\begin{table*}[t]
\begin{center}
\caption{The performance of TextDefense comparing to FGWS, DISP and TextShield.}
\SLJ{
	\scalebox{1}{\begin{tabular}{c c c c c c c c c c c c c c} 
		\toprule
        \multirow{2}{*}{Dataset}   & \multirow{2}{*}{Attack}    && \multicolumn{2}{c}{FGWS} && \multicolumn{2}{c}{DISP}    && \multicolumn{2}{c}{TextShield}  && \multicolumn{2}{c}{TextDefense}   \\
        \cline{4-5} \cline{7-8} \cline{10-11} \cline{13-14}
        &&& AUC   & F1    && AUC  & F1  && AUC  & F1   && AUC   & F1    \\
		\hline
		\multirow{3}{*}{Amazon}
        & TextBugger    && 0.7030    & 0.6952    &&  0.8893	& 0.8364 && \textbf{0.9846}    & 0.9680    && 0.9681 & 0.9249    \\
        & TextFooler    && 0.8311    & 0.8210    &&  0.8502	& 0.8150 && 0.9538    & 0.9437    && \textbf{0.9754} & 0.9333    \\
		& PWWS          && 0.8288    & 0.8101    &&  0.8381 & 0.8067 && 0.8490    & 0.8464    && \textbf{0.9589} & 0.9333 \\
		\hline
		\multirow{3}{*}{Yelp}
        & TextBugger    && 0.7647    & 0.7575    &&  0.8497	& 0.8174 && \textbf{0.9761}	  & 0.9428    && 0.9555    & 0.9172\\
        & TextFooler    && 0.7753    & 0.7637    &&  0.7976	& 0.7919 && 0.9586    & 0.9170    && \textbf{0.9609}    & 0.9207    \\
		& PWWS          && 0.8891    & 0.8790    &&  0.8023	& 0.7903 && 0.8975    & 0.8601    && \textbf{0.9536}    & 0.9110 \\
		\hline
		\multirow{3}{*}{Twitter}
        & TextBugger    && 0.7889    & 0.7454    &&  0.7872	& 0.7112 && 0.8463    & 0.7848    && \textbf{0.8807}    & 0.8372    \\
        & TextFooler    && 0.7031    & 0.6351    &&  0.6953	& 0.6410 && 0.8243    & 0.7713    && \textbf{0.9298}    & 0.8882     \\
		& PWWS          && 0.7224    & 0.6584    &&  0.6492	& 0.5858 && 0.7226    & 0.6881    && \textbf{0.8352}    & 0.7609 \\
		\bottomrule
	\end{tabular}}}
\label{table:disp_compare}
\end{center}
\vspace{-6mm}
\end{table*}

\section{Evaluation of Effectiveness}
\subsection{Detection Performance}\label{sec:clean_performance}
In this section, we evaluate the detection performance of TextDefense with two model architectures, three attack methods, and five binary classification datasets. 
We also evaluate its performance on multi-class classification datasets.
We first use several local adversarial texts and their original texts to determine the threshold with the highest F1 score.
In Table \ref{table:performance}, we report the ASR of each attack method on different models, the TPR and FPR under the predetermined threshold (TH), the semantic similarity between adversarial text and its original one, and the AUC of TextDefense of the five binary classification datasets. 
We also provide a detailed analysis of the trade-off between TPR and FPR in Appendix \ref{appendix:trade-off}.
In Table \ref{table:multiclass}, we report the performance of TextDefense of the two multi-class classification datasets.
Specifically, the threshold is pre-determined using 100 adversarial texts generated by TextBugger and TextFooler and their clean texts under the highest F1 score.

We can see from Table \ref{table:performance} that TextDefense achieves high AUC across different datasets, models, and attack methods. 
In the Amazon dataset, we can achieve an average AUC of 0.9698, which indicates that TextDefense can detect almost all adversarial examples with nearly no false positives.
\SLJ{At this time, we can set a threshold of 0.06, and the TPR and FPR are 0.96 and 0.12, respectively.
We can also see that the semantic similarity between adversarial examples generated by TextBugger and original texts is 0.8760.
This indicates that a few texts have been modified significantly.}
We find that the performance of these attack methods is different across the datasets, but TextDefense still performs impressively. 
For example, TextBugger has achieved an 85.65\% ASR on the BERT model trained on the Amazon dataset, and our TextDefense has an AUC of 0.9750. 
TextFooler and PWWS have achieved much higher ASR, but TextDefense can still detect adversarial texts with high accuracy and low FPR. 
\SLJ{The semantic similarity between original texts and adversarial examples generated by TextFooler and PWWS is slightly higher than that of TextBugger.
This is because character-level attack leads to unknown words, yet synonym replacement can retain text's semantics.
}
On the Twitter and Jigsaw datasets, the attack algorithms have achieved relatively low ASR compared to the sentimental analysis datasets, which we speculate that the model is more robust or that these datasets are more difficult to attack. 
The performance of TextDefense has slightly degraded in these two datasets. 
\SLJ{For the toxic comment datasets, the semantic similarity of texts attacked from TextBugger is higher than that from TextFooler and PWWS. 
This is because word-level attacks usually change the toxic words into non-toxic ones, which significantly reverse the toxicity of the text.} 
We speculate that the performance degradation is caused by replacing abusive words with other innocuous words, which would make the adversarial examples less abusive.
We defer the detailed analysis on the toxic comment datasets in Appendix \ref{case_twitter} and \ref{appendix:pwws_rule}.
Due to the space limit, we put the performance of TextDefense on other attack methods in Appendix \ref{appendix:other_attack}, where the results indicate that TextDefense can detect the adversarial examples with high AUC in all existing attacks. 
Therefore, we can infer that TextDefense is a universal detection method capable of both existing adversarial attacks and unknown adversarial attacks.

From Table \ref{table:performance}, we can observe similar results on LSTM models.
We can see that the LSTM models have a lower clean accuracy but higher ASR, which means that the LSTM model is less robust than the BERT model. 
However, TextDefense can still achieve a high AUC, which indicates that TextDefense can be extended to RNN-based models. 
Unlike the BERT model, the LSTM model fails to learn enough information from the IMDb dataset, resulting in the low accuracy of the IMDb model. 
This also indirectly leads to the performance degradation of TextDefense, where the average AUC is only 0.8142. 

\SLJ{Table \ref{table:multiclass} shows that TextDefense can also perform well on the two multi-class classification datasets. Specifically, TextDefense performs impressively on the AgNews dataset and is slightly weak on the Emotion dataset. This is because there exist different classes with similar semantic labels. For instance, the `love' and `joy' in the Emotion dataset are the vulnerable classes to attack. However, in AgNews, the texts under different labels have obvious semantic meanings.
}

\subsection{Method Comparison}\label{sec:comparison}
\SLJ{In this section, we compare the performance of TextDefense with the three methods mentioned in Sec. \ref{sec:baseline} and report AUC and F1 score under optimal threshold}
From Table \ref{table:disp_compare}, we can see that TextDefense outperforms the other three defense methods. 
\SLJ{We also compare the run-time and extra storage consumption in Appendix \ref{appendix:runtime} between TextDefense and the other three methods.} 
In the following, we give a detailed analysis of each compared method.

\paragraphbe{FGWS}
From Table \ref{table:disp_compare}, we can see that FGWS performs better on TextFooler and PWWS compared to TextBugger. 
For example, on the Amazon dataset, the F1 score of FGWS on TextFooler and PWWS is 0.13 higher than that on TextBugger. 
This means FGWS has a better detection ability on synonym substitution attacks. 
However, due to the intrinsic property of FGWS, it cannot locate the words subject to character-level attacks since the generated words have no synonyms.
Therefore, its detection capability on TextBugger is limited.
Nonetheless, we can see that the performance of TextDefense is not affected by the attack type, as the AUC is almost the same across the three attacks.

\paragraphbe{DISP} 
DISP is a method first to detect perturbed words and then restore them, which is not a pure detection method. 
Solely using the percentage of detected perturbed words in the input text cannot effectively distinguish adversarial examples which we have shown in Appendix \ref{appendix:disp}. 
Thus, we use the same way as FGWS to assess DISP, where if the change of prediction score exceeds a threshold, the text is considered adversarial. 

From Table \ref{table:disp_compare}, we can observe that the AUC of DISP on TextBugger is higher than that on TextFooler and PWWS. 
This difference indicates that the detection ability of DISP is more accurate on character-level attacks than word-level attacks since TextFooler and PWWS are pure synonym substitution methods. 
This is because DISP uses a trained model to detect perturbed words. 
We have known that the character-level attack modifies a word into one that does not exist in the model's vocabulary and, thus, is easier to detect. 
However, the poor performance of DISP on TextFooler and PWWS indicates that it is not good on the synonym substitution attack.
This is because DISP's discriminator can hardly detect perturbations of synonyms.
Nonetheless, TextDefense does not rely on a trained discriminator and can detect both attacks from synonyms and out-of-vocabulary words.

\paragraphbe{TextShield}
The main part of TextShield is the adversarial translation module, which is an LSTM model as stated in \cite{li2020textshield}. 
Here, we employ a pre-trained T5 \cite{2020t5}, the SOTA translation model, to translate the adversarial example into its original form. 
The T5 model is fine-tuned using 10k adversarial examples generated by TextBugger and TextFooler on the Amazon dataset.
We report the AUC and the F1 score of TextShield.

Table \ref{table:disp_compare} shows that TextShield performs impressively on the attack of TextBugger. 
This result justifies that TextShield has learned enough knowledge from the attack of TextBugger. 
The performance of TextShield has slightly dropped on the TextFooler dataset. 
This is because the searching space of the synonym substitution attack is huge, and TextShield cannot elaborate on all synonyms.
Furthermore, the performance of TextShield drops significantly on the PWWS attack. 
Since we do not consider adversarial examples from PWWS in the training phase, it is reasonable that TextShield cannot perform well. 
We also test TextShield on other attack methods in Appendix \ref{appendix:other_attack}, and the result indicates that TextShield fails to detect adversarial examples from new attacks.
Therefore, TextShield cannot transfer to unknown attack types, which, thus, is not comprehensively practical.
Finally, TextShield is inefficient in detecting adversarial examples since it has to generate text using an NMT model, which takes much time.
Nonetheless, TextDefense requires no prior knowledge about the attack, which can generalize to all types of attacks. 
Thus, TextDefense significantly outperforms TextShield in the attacks, which TextShield does not see.

In summary, TextDefense can detect adversarial examples with high AUC from both character-level and word-level attacks.  
\SLJ{That is, TextDefense has no limitation on the types of attacks and can be applied to both existing attacks and unknown attacks.
}

\begin{figure*}[t]
	\centering
	\includegraphics[width=17cm]{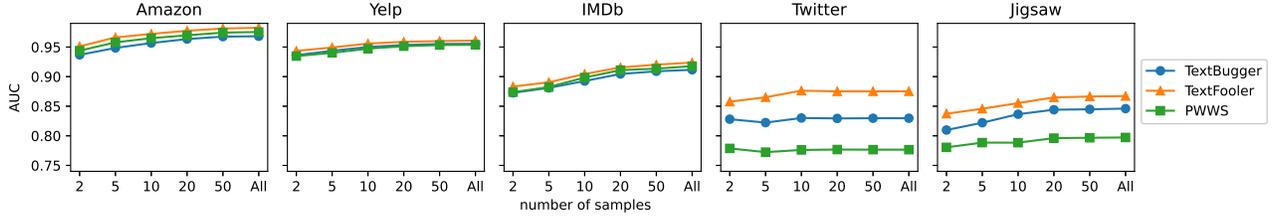}
	\vspace{-2mm}
	\caption{Sample size vs AUC of TextDefense.}
	\label{fig:sampling}
	\vspace{-4mm}
\end{figure*}

\begin{table}[h]
\begin{center}
\caption{The performance of TextDefense on dataset shift.}
	\scalebox{0.62}{\begin{tabular}{c c c c c c c c} 
		\toprule
        \multirow{2}{*}{Model}   & \multirow{2}{*}{Dataset}    & \multicolumn{2}{c}{TextBugger}    & \multicolumn{2}{c}{TextFooler}  & \multicolumn{2}{c}{PWWS}   \\
        \cline{3-8}
        && ASR   & AUC   & ASR    & AUC  & ASR    & AUC   \\
		\hline
		\multirow{2}{*}{Amazon}
        & \makecell{IMDb (90.4\%)} & 93.81\%    & 0.9305   & 96.68\% & 0.9378    & 96.24\%    & 0.9157 \\
        &  \makecell{Yelp (93.8\%)} & 92.96\%    & 0.9479   & 99.57\% & 0.9477    & 98.51\%    & 0.9352 \\
		\hline
		\multirow{2}{*}{Yelp}
        &  \makecell{Amazon (92.2\%)} & 90.24\%    & 0.9757   & 99.13\% & 0.9820    & 97.83\%    & 0.9819 \\
        &  \makecell{IMDb (89.2\%)} & 98.65\%    & 0.9264   & 100.00\% & 0.9265    & 100.00\%    & 0.9128 \\
		\hline
		\multirow{2}{*}{IMDb}
        &  \makecell{Amazon (89.2\%)} & 90.48\%    & 0.9394    & 99.57\%    & 0.9534   & 96.97\% & 0.9405 \\
        &  \makecell{Yelp (92.6\%)} & 91.14\%    & 0.8910    & 98.92\%    & 0.9065    & 98.92\%  & 0.9029 \\
		\bottomrule
	\end{tabular}}
\label{table:shift}
\end{center}
\vspace{-8mm}
\end{table}

\subsection{Evaluation of Transferability}\label{sec:transferability}
\paragraphbe{Dataset Shift}\label{sec:dataset_shift}
In the real world, there exists a scenario where a model (e.g., sentimental analysis model) trained on one dataset (e.g., Amazon dataset) is used to predict data from another dataset (e.g., Yelp dataset). 
The distribution of the two datasets may differ, whereas the target is the same. 
Previous defense methods directly detect the input text and do not rely on the target model, whereas TextDefense, which relies on the target model, may encounter the problem of out-of-distribution (OOD) examples. 
To justify that TextDefense can also detect adversarial OOD examples, we use three models from Sec. \ref{sec:clean_performance} which are trained on the Amazon, Yelp, and IMDb datasets, respectively, to predict the other two datasets and conduct adversarial attacks and detection. 
In Table \ref{table:shift}, we report the dataset used for training the target model (the Model column), the dataset used for generating adversarial examples on the target model (the Dataset column) together with their prediction accuracy, the ASR of the attack method and the AUC of TextDefense. 

From Table \ref{table:shift}, we can see that TextDefense has high AUC values under all circumstances.
Specifically, the Amazon model can achieve an accuracy of 93.8\% on the Yelp dataset, and the TextDefense achieves an average of 0.9436 AUC. 
The resulting AUC value is close to the performance on the Amazon dataset using the Amazon model.
Similar results can be found in other models with different datasets.
Therefore, we can conclude that TextDefense can defend text from another distribution but the same target. 

Previously in Table \ref{table:performance}, we can see that the detection of adversarial Yelp text in the Yelp model can achieve an average AUC value of 0.9660. 
However, in Table \ref{table:shift}, the performance of TextDefense on the adversarial Amazon texts using the Yelp model has a 0.9799 average AUC which is higher than 0.9660. 
Moreover, the ASR under the Yelp model is much higher than the Amazon model, which indicates that the Yelp model is less robust.
We hypothesize that the improved performance comes from the overfitting of the Yelp model, which we defer the detailed analysis in Sec. \ref{general}.
We can also see that the performance of TextDefense on the adversarial IMDb texts using the Yelp model has a drop compared to the performance on Amazon text. 
A similar observation can be found in the IMDb model. 
We speculate that the generated adversarial Amazon texts are closer to the decision boundary, which leads to the higher AUC performance on adversarial Amazon texts than on the adversarial Yelp texts on a Yelp model.
We defer the detailed analysis in Sec. \ref{sec:strong}. 

\paragraphbe{Transfer Attack}\label{sec:transfer}
Due to the intrinsic property that the capability of TextDefense relies on the performance of the target model, we are not sure whether TextDefense can detect transfer attacks from other models.
To verify that TextDefense can also defend transferred adversarial examples, we study its performance on transfer attacks in this section. 
We first train two target models using Amazon and Yelp datasets. 
Next, we train two surrogate models for each target model using samples from Amazon and Yelp that are disjoint with the target model training set.
Then, we use the surrogate model to generate adversarial examples and attack the target model. 
Finally, we perform defense on the target model and report the ASR of the transferred adversarial examples and the AUC of TextDefense.

\begin{table}[t]
\begin{center}
\caption{TextDefense's performance on transferred data.}
	\scalebox{0.7}{\begin{tabular}{c c c c c c c c} 
		\toprule
        \multirow{2}{*}{\makecell{Target\\ Model}}   & \multirow{2}{*}{\makecell{Surrogate\\ Model}}    & \multicolumn{2}{c}{TextBugger}    & \multicolumn{2}{c}{TextFooler}  & \multicolumn{2}{c}{PWWS}   \\
        \cline{3-8}
        && ASR   & AUC   & ASR    & AUC  & ASR    & AUC   \\
		\hline
		\multirow{2}{*}{Amazon}
        & Amazon  & 61.83\%    & 0.9399   & 78.04\% & 0.9105    & 69.08\%    & 0.9285 \\
        & Yelp  & 56.70\%    & 0.8628   & 67.63\% & 0.8129    & 65.36\%    & 0.8284 \\
		\hline
		\multirow{2}{*}{Yelp}
        & Amazon    & 50.42\%    & 0.9008   & 61.23\%   & 0.8835    & 55.30\%   & 0.8782 \\
        & Yelp      & 51.45\%    & 0.8680   & 57.47\%  & 0.8529    & 54.77\%  & 0.8923 \\
		\bottomrule
	\end{tabular}}
\label{table:transfer}
\end{center}
\vspace{-6mm}
\end{table}

The result from Table \ref{table:transfer} indicates that TextDefense can detect adversarial examples from transfer attacks. 
In the Amazon model, we can see that the adversarial Amazon examples generated from the surrogate Amazon model have better transferability than those generated from the surrogate Yelp model. 
We also observe that TextDefense performs better in the transfer attack from Amazon samples.
However, in the Yelp model, we find that the transferability of adversarial Yelp examples is slightly weaker than the transferability of adversarial Amazon examples, which contradicts the result from the Amazon model. 
This is probably because the trained Amazon model has better generalizability than the Yelp model, where the accuracy of the Amazon model on Yelp data is 97\%, and the accuracy of the Yelp model on Amazon data is 94\%.
We hypothesize that the performance of the TextDefense highly relies on the model's generalization. 
To justify the hypothesis, we study the relationship between the generalizability of the model and the performance of TextDefense in Sec. \ref{general}. 

\subsection{Influence of the Model Generalizability}\label{general}
In this section, we study how TextDefense performs in different model properties. 
One is the model's generalizability, which we have speculated in Sec. \ref{sec:transferability}. 
We also investigate whether TextDefense's capability comes from the model's generalizability. 
In order to quantify the generalizability, validation accuracy and loss are commonly used. 
The higher the validation accuracy or, the lower the validation loss, the better the model's generalizability. 

\begin{figure}[h]
	\centering
	\includegraphics[width=8cm]{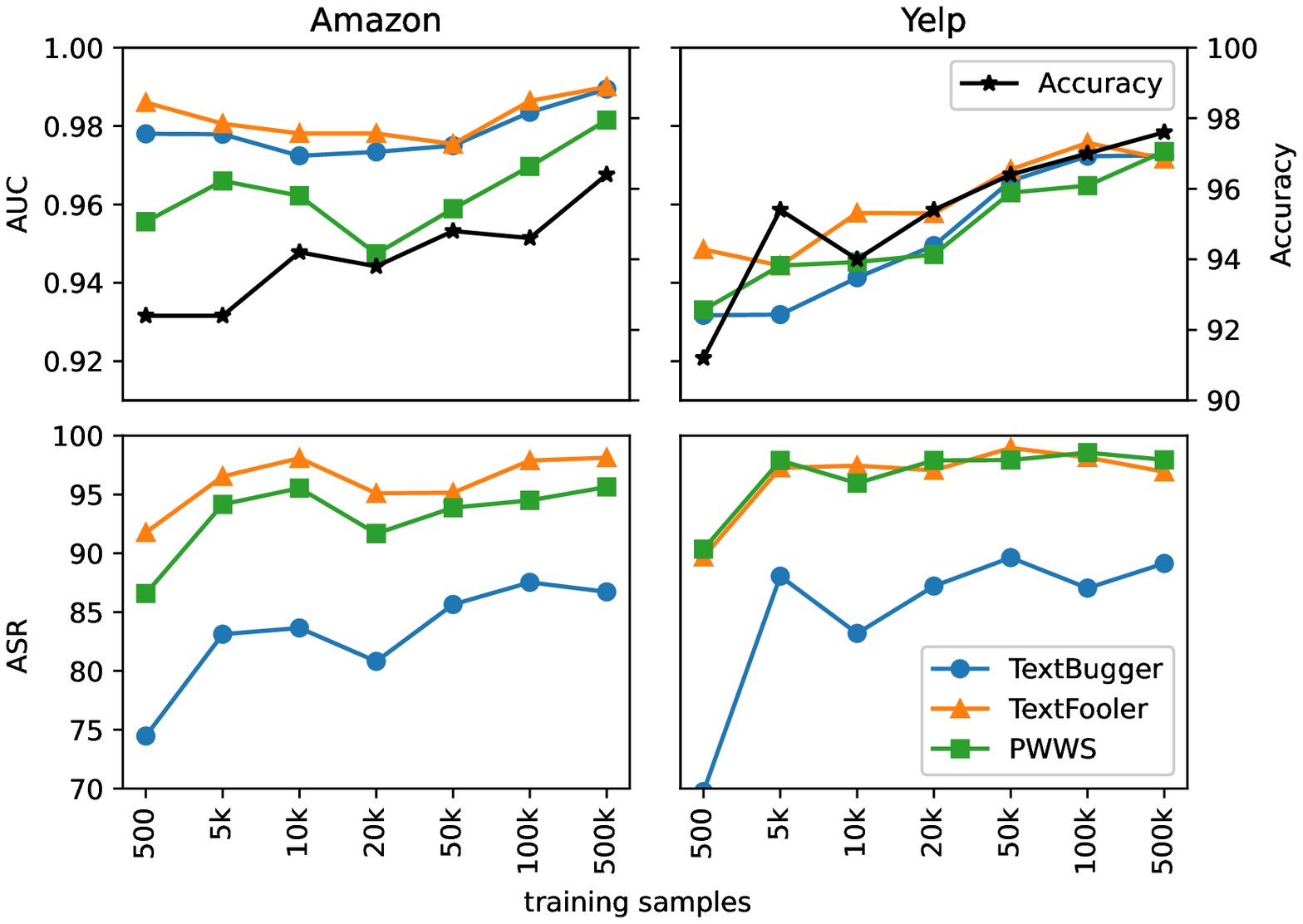}
	\vspace{-2mm}
	\caption{The relationship between the model's generalizability and the TextDefense's performance.}
	\label{fig:generalizability}
	\vspace{-2mm}
\end{figure}

\paragraphbe{Generalizability from Training Samples}
First, we study how the number of training samples influences a model's generalizability and further affects the performance of TextDefense. 
We train seven models for two datasets whose training samples range from 500 to 500k. 
The training process is stopped when it reaches the lowest validation loss.
Then, we use three attack methods to generate adversarial examples for each model. 
Finally, we report the AUC of TextDefense, the ASR of three attack methods, and the clean accuracy of each model in Fig. \ref{fig:generalizability}. 

From Fig. \ref{fig:generalizability}, we can observe that with the increasing number of training samples, the validation accuracy also increases. 
This indicates that the model trained using more samples has better generalizability than that trained with fewer samples. 
It can be seen that the higher the validation accuracy, the higher the AUC value of TextDefense. 
For example, in the Yelp model, when the training samples are 500, the model accuracy on the clean sample is 91.20\%. 
Meanwhile, the average AUC of TextDefense on three attack methods is 0.9378. When the training samples increase to 500k, the clean accuracy reaches 97.60\%, and the average AUC of TextDefense reaches 0.9725. 
Additionally, the increase in AUC is consistent with the increase in validation accuracy. 

From the ASR curve in Fig. \ref{fig:generalizability}, we can observe that the attack success rate increases when training samples increase. 
For example, in the Amazon dataset, the average ASR of the three attack methods is 84.27\% when the number of training samples is 500. 
However, the average ASR of the three methods increases to 94.86\% when the training samples increase to 500k. 
On the Yelp dataset, the average ASR increases from 83.26\% to 94.67\%. 
This indicates that the more training samples of the model, the less robust the model is. 
We conjecture that a model trained with more training samples will form a very complex decision boundary, resulting in more samples getting closer to the decision boundary. 
In the meantime, the model trained with fewer training samples is more robust against adversarial attacks due to the straightforward decision boundary. 
Nevertheless, the complex decision boundary also facilitates TextDefense's detection of adversarial examples. 
In Sec.\ref{sec:strong}, we also study the effect of the distance between decision boundary and adversarial example on the performance of TextDefense.

\begin{figure}[h]
	\centering
	\includegraphics[width=7cm]{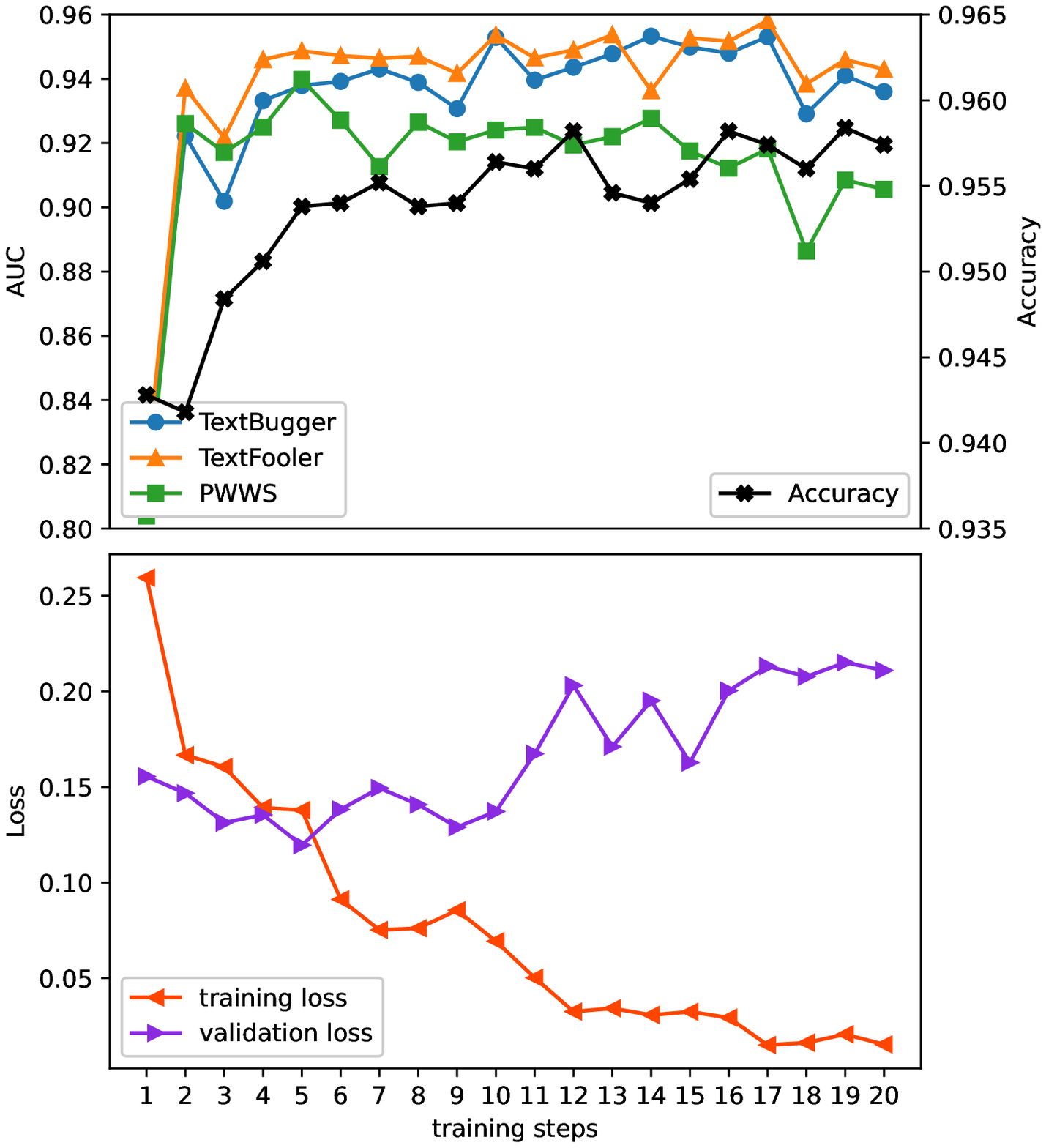}
	\vspace{-4mm}
	\caption{The relationship between the model's generalizability and the TextDefense's performance.}
	\label{fig:generalizability_steps}
	\vspace{-4mm}
\end{figure}

\paragraphbe{Generalizability from Training Steps}
In the previous experiment, we study the relationship between the performance of TextDefense and the generalizability coming from the number of training samples. 
The result indicates that the higher generalizability, the better the TextDefense performance. 
We also speculate that the complex decision boundary of the model, on the one hand, leads to a loss of robustness of the model but, on the other hand, improves the performance of TextDefense.
To further prove that the complexity of the decision boundary can facilitate TextDefense's performance, we study their relationship in this part.

As we know, an underfitted model has a straightforward decision boundary. 
At this time, the validation accuracy is low, and both training loss and validation loss are high. 
When the model is overfitted, the decision boundary would be overcomplex.
At this time, the validation accuracy is low, training loss is low, and validation loss is high. 
Therefore, to experiment on decision boundaries of varying complexity, we train the Yelp model continuously and record the model parameters intermittently until it overfits. 
Specifically, we train the model in four epochs divided into 20 steps, where one epoch contains five steps.
We record the model's training loss, validation loss, and validation accuracy for each model. Then, we use TextDefense to defend against three attacks and record their AUC. The results are shown in Fig. \ref{fig:generalizability_steps}.

From Fig. \ref{fig:generalizability_steps}, we can observe that the training loss continuously drops, and the validation loss drops until the fifth step and gradually increases afterward. 
We can also see that the validation accuracy increases rapidly in the first five steps and fluctuate afterward. 
We conjecture that the model has already formulated a complex decision boundary during the first epoch.
However, since we use the pre-trained model, we can see from the accuracy curve that the model does not overfit seriously.
That is, due to the regularization by the pre-training \cite{erhan2010does}, the model would not formulate an overcomplex decision boundary afterward. 
Subsequent training steps only modify the model's decision boundary slightly.
For the performance of TextDefense, we can see that the AUC increases steeply between the first and the fifth step and fluctuate afterward, which is consistent with the validation accuracy. 
Therefore, TextDefense gains its best performance after the first epoch of training.
Consequently, we can infer that the complexity of the model's decision boundary influences the performance of TextDefense.

\subsection{Evaluation on Different Models} 
\SLJ{In this section, we study the performance of TextDefense on different model architectures. 
We use the RoBERTa \cite{liu2019roberta}, DeBERTa \cite{he2020deberta}, ALBERT \cite{lan2019albert}, and BART \cite{lewis2020bart} to demonstrate the effectiveness of TextDefense, where RoBERTa, DeBERTa, and ALBERT are variants of the BERT model and BART is an encoder-decoder model. 
he results under the Amazon dataset is reported in Table \ref{table:architectures}.}

\SLJ{The table shows that TextDefense performs well on different transformer architectures. 
In addition, DeBERTa is more robust than other architectures. 
In the meanwhile, TextDefense performs best on the DeBERTa model. 
Moreover, BART, which has a different architecture, has impressive performance compared to the other three BERT variants. 
This indicates TextDefense is applicable to different transformer architectures.}

\begin{table}[h]
\centering
\caption{TextDefense on different transformer architectures.}
\scalebox{0.77}{\begin{tabular}{ccccccc}
\toprule
\multirow{2}{*}{\makecell{Model\\ Architecture}}        & \multicolumn{2}{c}{TextBugger} & \multicolumn{2}{c}{TextFooler} & \multicolumn{2}{c}{PWWS} \\
        \cline{2-7}
        & ASR            & AUC           & ASR            & AUC           & ASR         & AUC        \\
        \hline
RoBERTa & 78.95\%        & 0.9715        & 98.95\%        & 0.9760         & 95.58\%     & 0.9796     \\
DeBERTa & 70.75\%        & 0.9802        & 97.30\%        & 0.9900          & 91.49\%     & 0.9815     \\
ALBERT  & 86.29\%        & 0.9836        & 97.26\%        & 0.9862        & 93.88\%     & 0.9682    \\
BART    & 85.47\%        & 0.9842        & 98.95\%        & 0.9863        & 95.37\%     & 0.9704	\\
\bottomrule
\end{tabular}}\label{table:architectures}
\vspace{-4mm}
\end{table}

\subsection{Online NLP Classifier}
In this section, we deploy TextDefense together with online NLP classifiers. 
We select three pure model-based APIs: Symanto, NLPTown, and CardiffNLP. 
Other APIs containing post-processing of logits are not considered.
Since the slow return speed of online API calls, we use the transfer attack with the Amazon dataset to attack the three APIs, and the successfully attacked samples and their corresponding original samples are used to test TextDefense. 
We report the accuracy of three APIs, the ASR of the transfer attack, and the performance of TextDefense in Table \ref{table:online}.

 \begin{table}[h]
 \begin{center}
 \caption{The performance of TextDefense on online NLP classifers.}
 	\scalebox{0.7}{\begin{tabular}{c c c c c c c c} 
 		\toprule
         \multirow{2}{*}{API} & \multirow{2}{*}{Accuracy} & \multicolumn{2}{c}{TextBugger} & \multicolumn{2}{c}{TextFooler} & \multicolumn{2}{c}{PWWS}\\
         \cline{3-8}
         & & ASR & AUC & ASR & AUC & ASR & AUC \\
 		\hline
         Symanto & 89.81\% & 0.1991 & 0.8433 & 0.2081 & 0.8406 & 0.3839 & 0.8244\\
         NLPTown & 96.97\% & 0.2172 & 0.8794 & 0.2081 & 0.8906 & 0.3080 & 0.9065\\
         CardiffNLP & 93.43\% & 0.1490 & 0.8779 & 0.1253 & 0.9471 & 0.1793 & 0.9176\\
 		\bottomrule
 	\end{tabular}}
 \label{table:online}
 \end{center}
 \vspace{-4mm}
 \end{table}

\begin{table*}[t]
\centering
\caption{\SLJ{The attack success rate of adaptive attack using nine attack methods and the empirical robustness of TextDefense.}}
\SLJ{
\scalebox{0.73}{
\begin{tabular}{cccccccccccccc}
\toprule
Threshold & TPR  & FPR  & F1     & TextBugger & TextFooler & PWWS    & BAE     & DeepWordBug & Pruthi & CheckList & BERT-Attack & A2T    & \textit{Robustness} \\
\hline
0.04      & 1.00 & 0.17 & 0.9195 & 7.50\%     & 11.04\%    & 7.92\%  & 3.12\%  & 2.29\%      & 1.04\% & 0.21\%    & 8.33\%      & 1.04\% & 75.00\%    \\
0.05      & 0.98 & 0.14 & 0.9268 & 14.79\%    & 21.04\%    & 16.25\% & 5.21\%  & 5.00\%      & 2.29\% & 0.21\%    & 14.58\%     & 3.12\% & 64.60\%    \\
0.06      & 0.97 & 0.11 & 0.9317 & 26.25\%    & 30.83\%    & 23.33\% & 10.21\% & 11.46\%     & 3.54\% & 0.21\%    & 20.00\%     & 5.83\% & 53.40\%    \\
0.07      & 0.95 & 0.09 & 0.9318 & 34.17\%    & 36.67\%    & 31.67\% & 13.12\% & 15.00\%     & 3.96\% & 0.21\%    & 23.96\%   & 8.96\%   & 45.00\%   \\
/ & / & / & / & 69.38\% & 82.92\% & 81.25\% & 37.29\% & 62.29\% & 8.96\% & 0.85\% & 61.25\% & 29.79\% & 8.20\%\\
\bottomrule
\end{tabular}}}\label{table:robustness}
\vspace{-4mm}
\end{table*}

From Table \ref{table:online}, we can see that TextDefense can perform well in the online NLP. 
NLPTown has the highest accuracy among all APIs, and TextDefense performs well on NLPTown, although we do not know its training dataset. 
CardiffNLP has slightly lower accuracy on the Amazon dataset, yet TextDefense performs significantly better on TextFooler and PWWS attacks.
Symanto's accuracy is only 89.81\%, and TextDefense can still detect adversarial examples, but the performance is slightly weaker.
In conclusion, TextDefense can detect character-level and word-level attacks in these complete black-box settings. 
Hence, TextDefense can be integrated with MLaaS to enhance its robustness.

\begin{table}[h]
\centering
\caption{Comparison with adversarial training.}
\SLJ{
\scalebox{0.73}{\begin{tabular}{ccccccc}
\toprule
            & \multicolumn{2}{c}{TextBugger} & \multicolumn{2}{c}{TextFooler} & \multicolumn{2}{c}{PWWS} \\
            \cline{2-7}
            & ASR        & PW        & ASR        & PW        & ASR         & PW         \\
            \hline
Original         & 83.90\%    & 27.88\%   & 94.70\%    & 9.92\%    & 92.16\%     & 8.46\%     \\
DNE         & 75.37\%    & 32.17\%   & 82.87\%    & 9.95\%    & 76.23\%     & 7.48\%     \\
A2T         & 66.52\%    & 28.96\%   & 86.57\%    & 15.12\%   & 85.50\%     & 9.97\%     \\
TextDefense & \textbf{54.24\%}    & \textbf{42.28\%}   & \textbf{59.53\%}    & \textbf{15.36\%}   & \textbf{45.76\%}     & \textbf{12.17\%}   \\
\bottomrule
\end{tabular}}}\label{tab:AT}
\vspace{-4mm}
\end{table}

\begin{table*}[t]
\centering
\caption{TextDefense under adversarial examples with high confidence.}
\scalebox{0.85}{\begin{tabular}{ccccccccccccccc}
\toprule
\multirow{2}{*}{Target Score}   & \multicolumn{4}{c}{TextBugger}    && \multicolumn{4}{c}{TextFooler}     && \multicolumn{4}{c}{PWWS}          \\
\cline{2-5} \cline{7-10} \cline{12-15}   & ASR     & PW & AUC & LA    && ASR    & PW & AUC & LA    && ASR     & PW & AUC & LA   \\
   \hline
0.20 & 66.94\% & 35.05\%        & 0.8853 & 0.4601 && 93.02\% & 13.12\%        & 0.9295 & 0.4150  && 86.65\% & 10.93\%        & 0.9104 & 0.4384 \\
0.10 & 56.91\% & 37.99\%        & 0.7979 & 0.3464 && 90.85\% & 15.95\%        & 0.8601 & 0.3669 && 81.10\% & 12.96\%        & 0.8398 & 0.4110 \\
0.05 & 45.55\% & 39.50\%        & 0.6941 & 0.3022 && 87.04\% & 19.48\%        & 0.7833 & 0.3000 && 71.86\% & 14.40\%        & 0.7661 & 0.3887 \\
0.02 & 30.10\% & 40.99\%        & 0.6013 & 0.2215 && 76.16\% & 23.26\%        & 0.6624 & 0.2175 && 57.37\% & 16.49\%        & 0.6710 & 0.3310 \\
\bottomrule
\end{tabular}}\label{table:strong}
\vspace{-4mm}
\end{table*}

\section{Evaluation of Robustness}

\subsection{Adaptive Attack and Comparison with Adversarial Training}
\paragraphbe{Adaptive Attack} 
In this section, we evaluate the robustness of the TextDefense against adaptive attacks. 
\SLJ{We consider an attacker who is aware of the existence of TextDefense and intend to evade the defense. 
Meanwhile, the attacker also needs retain the readability and grammaticality of the generated text, so the attacker has to restrict the allowed perturbed words to 20\% of text words. 
As for defender, we should first determine the threshold for TextDefense based on our local samples. 
Therefore, we generate a total of 1000 local adversarial examples with TextBugger and TextFooler, and determine the threshold according to F1 scores, from which we obtain the threshold of approximately 0.04 to 0.07.
The FPR, TPR, and F1 scores of TextDefense on local samples are shown in Table \ref{table:robustness}
Then, we use nine different attacks to perform the adaptive attack and report the attack success rates of each attack method on different thresholds. 
We also provide the empirical robustness of TextDefense in the last column, which is calculated as the percentage of text that failed to be attacked by all nine methods.
} 

\SLJ{
It can be seen from the Table \ref{table:robustness} that the attack success rates have dropped to 1/2 to 1/4 of their original level when we set a threshold of 0.07. 
Moreover, when we set a threshold of 0.04, the attack success rates have dropped to 1/8 to 1/30 of their original level. 
The table also shows that when the threshold is set to 0.04, we can achieve a 1.00 TPR and 0.17 FPR on local adversarial and clean texts. 
This indicates that all locally generated adversarial examples can be detected, albeit with slightly higher false positives on clean samples. 
In the meanwhile, all nine attack methods achieve the lowest ASR.
Then, TextDefense can achieve 75\% empirical robustness, which means TextDefense is robust against 75\% of the input samples under the current attacks. 
When the threshold is set to 0.07, we can get a more appropriate TPR and FPR of 0.95 and 0.09, respectively.
At this time, the ASR of these methods is higher. 
However, TextDefense can still achieve 45\% empirical robustness. 
Empirically, if one wants a stronger protection capability, one may choose 0.04; and if one wants a lower false positive rate, one may choose 0.07.
}

\paragraphbe{Comparison with Adversarial Training}
In addition, we compare TextDefense with two SOTA adversarial training methods, DNE \cite{zhou2021defense} and A2T \cite{yoo-qi-2021-towards-improving}.
DNE is an adversarial training method combined with randomized smoothing to defend against synonym substitution attacks. 
A2T is another new and cheaper word substitution attack optimized for adversarial training. 
To align the metric of detection methods and adversarial training, we use the ASR after the model is protected. 
For the detection method, we use the ASR of adaptive attack under the threshold of 0.06, which we have illustrated in Table \ref{table:robustness}. 
For the adversarial training method, we directly use the ASR of the adversarially trained model.

From Table \ref{tab:AT}, we can see that the ASR of the adversarially trained models is lower than the ASR of the original trained model. 
This demonstrates that DNE and A2T have indeed improved the model's robustness from the ASR perspective. 
We then attack the model protected by TextDefense, and the average ASR is 53.18\%, much lower than the adversarially trained models.
Moreover, the percentage of perturbed words increases dramatically on the model protected by TextDefense. 
Therefore, TextDefense is more robust against adaptive attacks, which adversarial training usually cannot defend.

\subsection{Robustness against Adversarial Examples with High Confidence}\label{sec:strong}
Aside from the influence of the model, we also study the properties of adversarial examples affect TextDefense's performance. 
In previous experiments, our attack goal is to achieve an untargeted attack with less perturbed words; that is, the output logit for the ground truth label goes below 0.5. 
Therefore, most adversarial examples lie close to the decision boundary, which facilitates the detection. 
In this part, we force the output logit for the ground truth label to go below a target score. 
We use 0.20, 0.10, 0.05, and 0.02 as the target score to reveal the adversarial examples' influence on the performance of TextDefense.
Then, we perform adversarial attacks using the three attack methods with four target scores and report the ASR, PW, AUC of the TextDefense, and the LA of the generated adversarial examples in Table \ref{table:strong}. 

Table \ref{table:strong} shows that as we decrease the target score, the ASR decreases, the percentage of perturbed words increases, the AUC of TextDefense decreases, and the linguistic acceptability decreases for all the three attack methods. 
When the target score is 0.2, TextDefense can detect adversarial examples with high AUC. 
However, when the target score drops to 0.02, TextDefense has an average AUC of 0.6449, which is almost unusable. 
Nonetheless, at this time, the ASR of TextBugger drops significantly from 66.94\% to 30.10\%. 
Moreover, the linguistic acceptability of adversarial examples generated using TextBugger drops from 0.4601 to 0.2215. 
This indicates that only 22\% of generated examples are acceptable. 
Furthermore, at this time, 41\% of words are perturbed, which are no longer imperceptible to humans, thus violating the definition of adversarial examples.

For TextFooler and PWWS, similar observations can be discovered. 
However, their ASR remains relatively high, and PW also remains low. 
That is, TextFooler and PWWS are stronger attacks compared to TextBugger.
Thus, we suspect that stronger attacks like TextFooler and PWWS with high confidence may bypass TextDefense, and we defer the empirical study to Appendix \ref{case_tf}.

\section{Discussion}

In this section, we discuss the relationship between TextDefense and Randomized Smoothing and the challenges of TextDefense.

\subsection{Relationship with Randomized Smoothing}
Randomized smoothing \cite{cohen2019certified} has been proposed to guarantee that the smoothed classifier is robust around a sample $x$ within the $l_2$ radius $R$. 
To calculate the certified radius $R$ for the sample $x$, the randomized smoothing method perturbs $x$ by isotropic Gaussian noise.
What our approach and randomized smoothing in common is that we both perturb the input samples, whereas randomized smoothing perturbs inputs from the vector space by Gaussian noise and TextDefense perturbs inputs from the word space. 
However, due to the discretion of the language space, perturbing the vector representation of a text by Gaussian noise usually does not result in a corresponding text with physical meaning. 
Therefore, such a certified robustness radius is unrealistic for an NLP model.
Additionally, the smoothed classifier has to compute a plethora of perturbed inputs, which requires heavy computational resources.

Moreover, the application of randomized smoothing is currently limited to constructing a smoothed classifier with certified robustness.
Nevertheless, the constructed classifier does not help mitigate the model's vulnerability, and it can still be attacked by stronger attack algorithms successfully \cite{cohen2019certified, feng2020regularized}. 

\subsection{Challenges}
In previous experiments, we have shown that TextDefense is promising to be deployed in a real-world scenario. However, TextDefense has to query the model multiple times. 
With a sample size of 2, TextDefense still has to query the model three times which will triple the computation consumption.
If the size of the deployed model size is relatively small, then using TextDefense is acceptable. 
Whenever the model's size is huge, using TextDefense may bring greater pressure. 
In this case, we recommend training a small surrogate model to reduce resource consumption as we have known in Sec. \ref{sec:transfer} that transferred attack can also be defended by TextDefense.

\subsection{Limitations}
We have seen from previous sections that there exist several false negatives (i.e., low true positive rate) that TextDefense cannot detect. 
For instance, several adversarial examples from the Twitter and Jigsaw datasets cannot be detected.
By inspecting these false negatives, we can understand the rule of attacking toxic and non-toxic comments.
For example, in the adversarial Jigsaw texts attacked by PWWS, the toxic words in toxic comments like `stupid', `shit', and `suck' are commonly replaced by `stunned', `shop', and `draw', respectively.
These replaced words are no longer toxic, and these examples are predicted to the non-toxic label.  
On the contrary, the non-toxic words like `made', `know' and `mean' are replaced by `shit', `fuck', and `bastardly', respectively, which are all toxic words.
These `adversarial' examples are still considered adversarial and failed to be detected by TextDefense. 
However, their true label should be reversed. 
The adversarial examples generated using TextFooler has similar pattern.  
For more details, please refer to Appendix \ref{appendix:pwws_rule} and \ref{case_twitter}.

\section{Conclusion}
\label{sec:conclusion}

In this work, we propose TextDefense, the first universal framework for detecting adversarial examples in NLP. 
TextDefense uses the comprehensibility of the target model on the input text to discriminate the adversarial examples, that is, the information entropy contained in the text. 
Our evaluation of five datasets and multiple attack methods demonstrates that TextDefense can detect adversarial examples with high AUC and is robust under different circumstances. 
We also show that TextDefense outperforms the SOTA detection and adversarial training methods. 
With TextDefense, we can further understand how NLP models work in learning the knowledge and apprehend why adversarial examples exist in NLP.
%
%
%
%
%

\bibliographystyle{plain}
\bibliography{bib.bib}

\begin{thebibliography}{10}

\bibitem{alzantot2018generating}
Moustafa Alzantot, Yash Sharma, Ahmed Elgohary, Bo-Jhang Ho, Mani Srivastava,
  and Kai-Wei Chang.
\newblock Generating natural language adversarial examples.
\newblock In {\em Proceedings of the 2018 Conference on Empirical Methods in
  Natural Language Processing}, pages 2890--2896, 2018.

\bibitem{Balunovic2020Adversarial}
Mislav Balunovic and Martin Vechev.
\newblock Adversarial training and provable defenses: Bridging the gap.
\newblock In {\em International Conference on Learning Representations}, 2020.

\bibitem{cohen2019certified}
Jeremy Cohen, Elan Rosenfeld, and Zico Kolter.
\newblock Certified adversarial robustness via randomized smoothing.
\newblock In {\em International Conference on Machine Learning}, pages
  1310--1320. PMLR, 2019.

\bibitem{Devlin2019}
Jacob Devlin, Ming-Wei Chang, Kenton Lee, and Kristina Toutanova.
\newblock {BERT}: Pre-training of deep bidirectional transformers for language
  understanding.
\newblock In {\em Proceedings of the 2019 Conference of the North {A}merican
  Chapter of the Association for Computational Linguistics: Human Language
  Technologies, Volume 1 (Long and Short Papers)}, pages 4171--4186,
  Minneapolis, Minnesota, June 2019. Association for Computational Linguistics.

\bibitem{du2021cert}
Tianyu Du, Shouling Ji, Lujia Shen, Yao Zhang, Jinfeng Li, Jie Shi, Chengfang
  Fang, Jianwei Yin, Raheem Beyah, and Ting Wang.
\newblock Cert-rnn: Towards certifying the robustness of recurrent neural
  networks.
\newblock In {\em Proceedings of the 2021 ACM SIGSAC Conference on Computer and
  Communications Security}, pages 516--534, 2021.

\bibitem{du2021combating}
Xiaohu Du, Jie Yu, Shasha Li, Zibo Yi, Hai Liu, and Jun Ma.
\newblock Combating word-level adversarial text with robust adversarial
  training.
\newblock In {\em 2021 International Joint Conference on Neural Networks
  (IJCNN)}, pages 1--8. IEEE, 2021.

\bibitem{erhan2010does}
Dumitru Erhan, Aaron Courville, Yoshua Bengio, and Pascal Vincent.
\newblock Why does unsupervised pre-training help deep learning?
\newblock In {\em Proceedings of the thirteenth international conference on
  artificial intelligence and statistics}, pages 201--208. JMLR Workshop and
  Conference Proceedings, 2010.

\bibitem{fan2021adversarial}
Jiameng Fan and Wenchao Li.
\newblock Adversarial training and provable robustness: A tale of two
  objectives.
\newblock In {\em Proceedings of the AAAI Conference on Artificial
  Intelligence}, volume~35, pages 7367--7376, 2021.

\bibitem{feng2020regularized}
Huijie Feng, Chunpeng Wu, Guoyang Chen, Weifeng Zhang, and Yang Ning.
\newblock Regularized training and tight certification for randomized smoothed
  classifier with provable robustness.
\newblock In {\em Proceedings of the AAAI Conference on Artificial
  Intelligence}, volume~34, pages 3858--3865, 2020.

\bibitem{founta2018large}
Antigoni-Maria Founta, Constantinos Djouvas, Despoina Chatzakou, Ilias
  Leontiadis, Jeremy Blackburn, Gianluca Stringhini, Athena Vakali, Michael
  Sirivianos, and Nicolas Kourtellis.
\newblock Large scale crowdsourcing and characterization of twitter abusive
  behavior.
\newblock In {\em 11th International Conference on Web and Social Media, ICWSM
  2018}. AAAI Press, 2018.

\bibitem{gao2018black}
Ji~Gao, Jack Lanchantin, Mary~Lou Soffa, and Yanjun Qi.
\newblock Black-box generation of adversarial text sequences to evade deep
  learning classifiers.
\newblock In {\em 2018 IEEE Security and Privacy Workshops (SPW)}, pages
  50--56, 2018.

\bibitem{garg2020bae}
Siddhant Garg and Goutham Ramakrishnan.
\newblock Bae: Bert-based adversarial examples for text classification.
\newblock In {\em Proceedings of the 2020 Conference on Empirical Methods in
  Natural Language Processing (EMNLP)}, pages 6174--6181, 2020.

\bibitem{goodfellow2014explaining}
Ian~J. Goodfellow, Jonathon Shlens, and Christian Szegedy.
\newblock Explaining and harnessing adversarial examples.
\newblock In Yoshua Bengio and Yann LeCun, editors, {\em 3rd International
  Conference on Learning Representations, {ICLR} 2015, San Diego, CA, USA, May
  7-9, 2015, Conference Track Proceedings}, 2015.

\bibitem{he2020deberta}
Pengcheng He, Xiaodong Liu, Jianfeng Gao, and Weizhu Chen.
\newblock Deberta: Decoding-enhanced bert with disentangled attention.
\newblock {\em arXiv preprint arXiv:2006.03654}, 2020.

\bibitem{LSTM}
Sepp Hochreiter and J\"{u}rgen Schmidhuber.
\newblock Long short-term memory.
\newblock {\em Neural Comput.}, 9(8):1735–1780, nov 1997.

\bibitem{jin2020bert}
Di~Jin, Zhijing Jin, Joey~Tianyi Zhou, and Peter Szolovits.
\newblock Is bert really robust? a strong baseline for natural language attack
  on text classification and entailment.
\newblock In {\em Proceedings of the AAAI conference on artificial
  intelligence}, volume~34, pages 8018--8025, 2020.

\bibitem{kale2020text}
Mihir Kale and Abhinav Rastogi.
\newblock Text-to-text pre-training for data-to-text tasks.
\newblock In {\em Proceedings of the 13th International Conference on Natural
  Language Generation}, pages 97--102, 2020.

\bibitem{ko2019popqorn}
Ching-Yun Ko, Zhaoyang Lyu, Lily Weng, Luca Daniel, Ngai Wong, and Dahua Lin.
\newblock Popqorn: Quantifying robustness of recurrent neural networks.
\newblock In {\em International Conference on Machine Learning}, pages
  3468--3477. PMLR, 2019.

\bibitem{kuleshov2018adversarial}
Volodymyr Kuleshov, Shantanu Thakoor, Tingfung Lau, and Stefano Ermon.
\newblock Adversarial examples for natural language classification problems,
  2018.

\bibitem{10.5555/3367243.3367425}
Nupur Kumari, Mayank Singh, Abhishek Sinha, Harshitha Machiraju, Balaji
  Krishnamurthy, and Vineeth~N. Balasubramanian.
\newblock Harnessing the vulnerability of latent layers in adversarially
  trained models.
\newblock In {\em Proceedings of the 28th International Joint Conference on
  Artificial Intelligence}, IJCAI'19, page 2779–2785. AAAI Press, 2019.

\bibitem{sym13030428}
Hyun Kwon and Jun Lee.
\newblock Diversity adversarial training against adversarial attack on deep
  neural networks.
\newblock {\em Symmetry}, 13(3), 2021.

\bibitem{lan2019albert}
Zhenzhong Lan, Mingda Chen, Sebastian Goodman, Kevin Gimpel, Piyush Sharma, and
  Radu Soricut.
\newblock Albert: A lite bert for self-supervised learning of language
  representations.
\newblock {\em arXiv preprint arXiv:1909.11942}, 2019.

\bibitem{lewis2020bart}
Mike Lewis, Yinhan Liu, Naman Goyal, Marjan Ghazvininejad, Abdelrahman Mohamed,
  Omer Levy, Veselin Stoyanov, and Luke Zettlemoyer.
\newblock Bart: Denoising sequence-to-sequence pre-training for natural
  language generation, translation, and comprehension.
\newblock In {\em Proceedings of the 58th Annual Meeting of the Association for
  Computational Linguistics}, pages 7871--7880, 2020.

\bibitem{li2020textshield}
Jinfeng Li, Tianyu Du, Shouling Ji, Rong Zhang, Quan Lu, Min Yang, and Ting
  Wang.
\newblock {\em TEXTSHIELD: Robust Text Classification Based on Multimodal
  Embedding and Neural Machine Translation}.
\newblock USENIX Association, USA, 2020.

\bibitem{li2018textbugger}
Jinfeng Li, Shouling Ji, Tianyu Du, Bo~Li, and Ting Wang.
\newblock Textbugger: Generating adversarial text against real-world
  applications.
\newblock In {\em 26th Annual Network and Distributed System Security
  Symposium, {NDSS} 2019, San Diego, California, USA, February 24-27, 2019}.
  The Internet Society, 2019.

\bibitem{li2020bert}
Linyang Li, Ruotian Ma, Qipeng Guo, Xiangyang Xue, and Xipeng Qiu.
\newblock Bert-attack: Adversarial attack against bert using bert.
\newblock In {\em Proceedings of the 2020 Conference on Empirical Methods in
  Natural Language Processing (EMNLP)}, pages 6193--6202, 2020.

\bibitem{liu2019roberta}
Yinhan Liu, Myle Ott, Naman Goyal, Jingfei Du, Mandar Joshi, Danqi Chen, Omer
  Levy, Mike Lewis, Luke Zettlemoyer, and Veselin Stoyanov.
\newblock Roberta: A robustly optimized bert pretraining approach.
\newblock {\em arXiv preprint arXiv:1907.11692}, 2019.

\bibitem{madry2018towards}
Aleksander Madry, Aleksandar Makelov, Ludwig Schmidt, Dimitris Tsipras, and
  Adrian Vladu.
\newblock Towards deep learning models resistant to adversarial attacks.
\newblock In {\em International Conference on Learning Representations}, 2018.

\bibitem{mozes2021frequency}
Maximilian Mozes, Pontus Stenetorp, Bennett Kleinberg, and Lewis Griffin.
\newblock Frequency-guided word substitutions for detecting textual adversarial
  examples.
\newblock In {\em Proceedings of the 16th Conference of the European Chapter of
  the Association for Computational Linguistics: Main Volume}, pages 171--186,
  2021.

\bibitem{papernot2016crafting}
Nicolas Papernot, Patrick McDaniel, Ananthram Swami, and Richard Harang.
\newblock Crafting adversarial input sequences for recurrent neural networks.
\newblock In {\em MILCOM 2016-2016 IEEE Military Communications Conference},
  pages 49--54. IEEE, 2016.

\bibitem{pruthi2019combating}
Danish Pruthi, Bhuwan Dhingra, and Zachary~C Lipton.
\newblock Combating adversarial misspellings with robust word recognition.
\newblock In {\em Proceedings of the 57th Annual Meeting of the Association for
  Computational Linguistics}, pages 5582--5591, 2019.

\bibitem{2020t5}
Colin Raffel, Noam Shazeer, Adam Roberts, Katherine Lee, Sharan Narang, Michael
  Matena, Yanqi Zhou, Wei Li, and Peter~J. Liu.
\newblock Exploring the limits of transfer learning with a unified text-to-text
  transformer.
\newblock {\em Journal of Machine Learning Research}, 21(140):1--67, 2020.

\bibitem{ren2019generating}
Shuhuai Ren, Yihe Deng, Kun He, and Wanxiang Che.
\newblock Generating natural language adversarial examples through probability
  weighted word saliency.
\newblock In {\em Proceedings of the 57th annual meeting of the association for
  computational linguistics}, pages 1085--1097, 2019.

\bibitem{schuster2012japanese}
Mike Schuster and Kaisuke Nakajima.
\newblock Japanese and korean voice search.
\newblock In {\em 2012 IEEE international conference on acoustics, speech and
  signal processing (ICASSP)}, pages 5149--5152. IEEE, 2012.

\bibitem{tramer2018ensemble}
Florian Tramèr, Alexey Kurakin, Nicolas Papernot, Ian Goodfellow, Dan Boneh,
  and Patrick McDaniel.
\newblock Ensemble adversarial training: Attacks and defenses.
\newblock In {\em International Conference on Learning Representations}, 2018.

\bibitem{wang2019towards}
W.~Wang, R.~Wang, L.~Wang, Z.~Wang, and A.~Ye.
\newblock Towards a robust deep neural network against adversarial texts: A
  survey.
\newblock {\em IEEE Transactions on Knowledge \& Data Engineering}, (01):1--1,
  oct 5555.

\bibitem{pmlr-v161-wang21a}
Xiaosen Wang, Jin Hao, Yichen Yang, and Kun He.
\newblock Natural language adversarial defense through synonym encoding.
\newblock In Cassio de~Campos and Marloes~H. Maathuis, editors, {\em
  Proceedings of the Thirty-Seventh Conference on Uncertainty in Artificial
  Intelligence}, volume 161 of {\em Proceedings of Machine Learning Research},
  pages 823--833. PMLR, 27--30 Jul 2021.

\bibitem{wang2020defense}
Zhaoyang Wang and Hongtao Wang.
\newblock Defense of word-level adversarial attacks via random substitution
  encoding.
\newblock In {\em International Conference on Knowledge Science, Engineering
  and Management}, pages 312--324. Springer, 2020.

\bibitem{warstadt2019neural}
Alex Warstadt, Amanpreet Singh, and Samuel~R Bowman.
\newblock Neural network acceptability judgments.
\newblock {\em Transactions of the Association for Computational Linguistics},
  7:625--641, 2019.

\bibitem{xu2019lexicalat}
Jingjing Xu, Liang Zhao, Hanqi Yan, Qi~Zeng, Yun Liang, and Xu~Sun.
\newblock {L}exical{AT}: Lexical-based adversarial reinforcement training for
  robust sentiment classification.
\newblock In {\em Proceedings of the 2019 Conference on Empirical Methods in
  Natural Language Processing and the 9th International Joint Conference on
  Natural Language Processing (EMNLP-IJCNLP)}, pages 5518--5527, Hong Kong,
  China, November 2019. Association for Computational Linguistics.

\bibitem{yang2019xlnet}
Zhilin Yang, Zihang Dai, Yiming Yang, Jaime Carbonell, Russ~R Salakhutdinov,
  and Quoc~V Le.
\newblock Xlnet: Generalized autoregressive pretraining for language
  understanding.
\newblock {\em Advances in neural information processing systems}, 32, 2019.

\bibitem{yoo-qi-2021-towards-improving}
Jin~Yong Yoo and Yanjun Qi.
\newblock Towards improving adversarial training of {NLP} models.
\newblock In {\em Findings of the Association for Computational Linguistics:
  EMNLP 2021}, pages 945--956, Punta Cana, Dominican Republic, November 2021.
  Association for Computational Linguistics.

\bibitem{zhou2021defense}
Yi~Zhou, Xiaoqing Zheng, Cho-Jui Hsieh, Kai-Wei Chang, and Xuanjing Huang.
\newblock Defense against synonym substitution-based adversarial attacks via
  {D}irichlet neighborhood ensemble.
\newblock In {\em Proceedings of the 59th Annual Meeting of the Association for
  Computational Linguistics and the 11th International Joint Conference on
  Natural Language Processing (Volume 1: Long Papers)}, pages 5482--5492,
  Online, August 2021. Association for Computational Linguistics.

\bibitem{zhou2019learning}
Yichao Zhou, Jyun-Yu Jiang, Kai-Wei Chang, and Wei Wang.
\newblock Learning to discriminate perturbations for blocking adversarial
  attacks in text classification.
\newblock pages 4906--4915, 01 2019.

\end{thebibliography}
\appendix

\section{Evaluation of Factors}\label{factors}
\subsection{Gradient Method}\label{sec:gradient}
In designing our method, we use the WIR algorithm instead of the gradient-based method because the gradient is not representative of a tokenized word, as stated in Sec. \ref{sec:method}.
To justify this assumption, we replace the importance scores obtained by WIR with deletion by the gradient of words. 
That is, the $L_1$ norm of the partial derivative of the loss function concerning the embedding of the word calculated by
$s_i = \left\Vert\frac{\partial\mathcal{L}(x)}{\partial w_i}\right\Vert$.
The resulting AUC of TextDefense is shown in Table \ref{tab:gradient}.

\begin{table}[h]
\centering
\caption{TextDefense's performance using gradient method.}
\begin{tabular}{cccc}
\toprule
       & TextBugger & TextFooler & PWWS   \\
       \hline
Amazon & 0.1796     & 0.1741     & 0.1193 \\
Yelp   & 0.2325     & 0.2617     & 0.2189 \\
\bottomrule
\end{tabular}\label{tab:gradient}
\end{table}

Compared to the AUC of TextDefense using the WIR algorithm, as shown in Table \ref{table:performance}, the performance of the gradient method is much worse. 
This indicates that the gradient of a word cannot capture the real importance of the word. 
We speculate that 1) words are correlated with each other in the text, and thus, the gradient of a single word might not be representative of the importance of the word;
2) the plane of the loss function is smooth, which results in an insignificant influence on the gradient;
3) the gradient is relatively low due to that the generated examples are close to the local extreme value;
4) the gradient method is not suitable for our detection method.

\subsection{Sampling Method}\label{sample}
Since it takes more time to calculate the importance scores of all words in an input text, we use the sampling method to speed up the importance score calculation process, which calculates only a portion of words in the input text.
In this section, we study how the sampling technique affects the performance of TextDefense. 
We use five sample sizes, where the number of words sampled is 2, 5, 10, 20, and 50, together with all words.
The sampled words are used to calculate the entropy of the input text. 
\SLJ{The average text length of Amazon, Yelp, IMDb, Twitter, and Jigsaw is 88, 149, 228, 16, and 71, respectively. 
We report the AUC of each sampling method in Fig. \ref{fig:sampling}.
Note that, in the Twitter dataset, using a sample size of 20 or 50 has the same performance as using all words.
We also report the run-time per sample with sampling technique in Appendix \ref{appendix:runtime}.
}

From Fig. \ref{fig:sampling}, we can see that in all datasets, the AUC value increases as the sample size increases. 
However, the increasing trend is not significant. 
In the Amazon dataset, the AUC is 0.9439 when the sample size is 2. 
When we sample all words in the text, the AUC increases by 0.03 to 0.9754.
We can also see that the increasing trend is apparent when the sample size increases from 2 to 10 (e.g., 0.9439 to 0.9647), while there is a minor increase in AUC when the sample size is increased from 10 to all words (e.g., 0.9647 to 0.9744). 
Therefore, while ensuring TextDefense's performance, we can choose the sampling method with a sample size of 10 to reduce the computational overhead.
TextDefense has a slight drop in performance in the Twitter and Jigsaw models. 
However, the sampling method barely changes the performance of TextDefense in these two datasets. 

\SLJ{We also calculate the run-time for each query using the sampling technique in Appendix \ref{appendix:runtime}. The experimental results indicate that the sampling method can significantly reduce the computation overhead while keeping the performance unchanged.}

\subsection{Scoring Metric}\label{sec:scoring}
In Sec. \ref{sec:method}, we use the entropy to quantify the scale of the importance scores under the intuition of the information retained by each word.
However, other scoring metrics can be used to quantify such scale in statistics, for example, the mean value. 
We study nine scoring metrics and report their performance in TextDefense to find the best one to distinguish the adversarial examples from the original ones.
To be specific, we use mean value, standard deviation, range, midrange (mean value of maximum and minimum), the first quartile, the third quartile, quartile deviation (the difference between the third quartile and the first quartile), coefficient of variance (quotient of standard deviation and mean value), and entropy. 
We report the performance of TextDefense using nine scoring metrics in Table \ref{table:scoring}.

\begin{table}[h]
\centering
\caption{The performance of TextDefense using different scoring metrics.}
\SLJ{
\scalebox{0.9}{
\begin{tabular}{cccc}
\toprule
Metric                  & TextBugger & TextFooler & PWWS \\
\hline
Mean                    & 0.6807     & 0.6753     & 0.7509    \\
Standard Deviation      & 0.9067     & 0.9451     & 0.9459     \\
Range                   & 0.8888     & 0.9269     & 0.9289     \\
Midrange                & 0.5905     & 0.6245     & 0.7486     \\
First Quartile          & 0.2234     & 0.2138     & 0.2275     \\
Third Quartile          & 0.9007     & 0.8912     & 0.9119     \\
Quartile Deviation      & 0.9653     & 0.9786     & 0.9715     \\
Coefficient of Variance & 0.3478     & 0.3555     & 0.4027     \\
Entropy                 & \textbf{0.9681}     & \textbf{0.9826}     & \textbf{0.9754}    \\
\bottomrule
\end{tabular}}}\label{table:scoring}
\end{table}

From Table \ref{table:scoring}, we can see that the entropy performs best across all nine metrics with an average AUC of 0.9754. 
In addition, the quartile deviation also can be used in TextDefense, where the average AUC is 0.9718, which is only 0.0036 lower than using entropy.
However, solely using the first or third quartile cannot reach a high AUC performance.
From the results, we can infer that the lower importance scores in the clean and adversarial text are distributed similarly due to the worst performance of the first quartile. 
Meanwhile, the larger importance scores in the clean and adversarial texts differ according to the performance of the third quartile. 
Therefore, the above results justify our previous assumption that the clean text has relatively low scales or several words with high importance scores apart from which it has low scales.
Additionally, from the standard deviation and range performance, we know that the word importance scores in adversarial examples have significant disparity. 

\begin{table*}[t]
\begin{center}
\caption{\SLJ{The performance of TextDefense on different attack methods with different thresholds.}}
\SLJ{
\scalebox{0.67}{
\begin{tabular}{cccccccccccccccccccccc}
\toprule
          & \multicolumn{3}{c}{TextBugger} & \multicolumn{3}{c}{TextyFooler} & \multicolumn{3}{c}{PWWS}      & \multicolumn{3}{c}{DeepWordBug} & \multicolumn{3}{c}{IGA}       & \multicolumn{3}{c}{Kuleshov et al.} & \multicolumn{3}{c}{BAE}       \\
          \hline
threshold & TPR   & FPR  & F1              & TPR   & FPR   & F1              & TPR  & FPR  & F1              & TPR   & FPR   & F1              & TPR  & FPR  & F1              & TPR    & FPR    & F1                & TPR  & FPR  & F1              \\
\hline
0.01      & 1.00     & 0.39 & 0.8372          & 1.00     & 0.36  & 0.8466          & 1.00    & 0.37 & 0.843           & 1.00     & 0.41  & 0.8291          & 1.00    & 0.44 & 0.8212          & 1.00      & 0.34   & 0.8533            & 1.00    & 0.48 & 0.8049          \\
0.02      & 0.99  & 0.27 & 0.8785          & 1.00     & 0.25  & 0.8873          & 0.99 & 0.26 & 0.8811          & 0.99  & 0.28  & 0.8728          & 0.99 & 0.30  & 0.8613          & 0.99   & 0.22   & 0.8952            & 0.99 & 0.35 & 0.8454          \\
0.03      & 0.99  & 0.21 & 0.9001          & 0.99  & 0.19  & 0.9108          & 0.98 & 0.19 & 0.9027          & 0.99  & 0.22  & 0.8966          & 0.98 & 0.24 & 0.8847          & 0.98   & 0.16   & 0.9161            & 0.98 & 0.28 & 0.8676          \\
0.04      & 0.98  & 0.17 & \textbf{0.9108} & 0.98  & 0.15  & \underline{0.9192} & 0.97 & 0.16 & \underline{0.9091} & 0.98  & 0.18  & \textit{0.9091} & 0.97 & 0.19 & \underline{0.8961} & 0.98   & 0.13   & \underline{0.9283}   & 0.97 & 0.23 & \textit{0.8836} \\
0.05      & 0.94  & 0.13 & \textit{0.9086} & 0.97  & 0.12  & \textbf{0.9305} & 0.95 & 0.12 & \textit{0.9168} & 0.97  & 0.14  & \textbf{0.9194} & 0.96 & 0.15 & \textbf{0.9079} & 0.96   & 0.09   & \textbf{0.9364}   & 0.97 & 0.19 & \textbf{0.8992} \\
0.06      & 0.91  & 0.10  & \underline{0.9031} & 0.95  & 0.09  & \textit{0.9298} & 0.93 & 0.09 & \textbf{0.9170}  & 0.91  & 0.11  & \underline{0.9012} & 0.92 & 0.12 & \textit{0.9033} & 0.94   & 0.07   & \textit{0.9360}    & 0.91 & 0.15 & \underline{0.8819} \\
0.07      & 0.84  & 0.09 & 0.8743          & 0.91  & 0.08  & 0.9144          & 0.89 & 0.08 & 0.9057          & 0.84  & 0.09  & 0.8678          & 0.89 & 0.10  & 0.8939          & 0.91   & 0.05   & 0.9266            & 0.86 & 0.13 & 0.8640           \\
0.08      & 0.78  & 0.07 & 0.8438          & 0.86  & 0.06  & 0.8961          & 0.82 & 0.06 & 0.8701          & 0.77  & 0.07  & 0.8369          & 0.85 & 0.07 & 0.8814          & 0.85   & 0.04   & 0.9021            & 0.80  & 0.10  & 0.8402          \\
0.09      & 0.70   & 0.05 & 0.7994          & 0.79  & 0.04  & 0.8637          & 0.76 & 0.05 & 0.8393          & 0.69  & 0.05  & 0.7951          & 0.80  & 0.06 & 0.864           & 0.80    & 0.02   & 0.8776            & 0.73 & 0.09 & 0.8067         \\
\bottomrule
\end{tabular}}}\label{table:threshold}
\end{center}
\end{table*}

\begin{table*}[t]
\centering
\caption{\SLJ{The performance of TextDefense on different numbers of local adversarial examples (the best in bold, second-best in italic, and third-best in underline).}}
\SLJ{
\begin{tabular}{ccccccccccccc}
\toprule
\multirow{2}{*}{Threshold}     & \multicolumn{3}{c}{50}        & \multicolumn{3}{c}{100}       & \multicolumn{3}{c}{200}       & \multicolumn{3}{c}{500}       \\
     \cline{2-13}
     & TPR  & FPR  & F1              & TPR  & FPR  & F1              & TPR  & FPR  & F1              & TPR  & FPR  & F1              \\
     \hline
0.01 & 1.00 & 0.36 & 0.8475          & 1.00 & 0.37 & 0.8439          & 1.00 & 0.38 & 0.8403          & 1.00 & 0.39 & 0.8363          \\
0.02 & 1.00 & 0.34 & 0.8547          & 1.00 & 0.32 & 0.8621          & 1.00 & 0.30 & 0.8696          & 1.00 & 0.29 & 0.8742          \\
0.03 & 1.00 & 0.22 & 0.9009          & 1.00 & 0.24 & 0.8929          & 0.99 & 0.23 & 0.8944          & 0.99 & 0.23 & 0.8934          \\
0.04 & 1.00 & 0.14 & \textbf{0.9346} & 1.00 & 0.15 & \textbf{0.9302} & 0.99 & 0.16 & \textit{0.9234} & 0.99 & 0.16 & \textbf{0.9204} \\
0.05 & 0.98 & 0.12 & \textit{0.9333} & 0.97 & 0.13 & \textit{0.9238} & 0.97 & 0.13 & \textbf{0.9264} & 0.96 & 0.13 & \textit{0.9194} \\
0.06 & 0.96 & 0.12 & \underline{0.9231}    & 0.96 & 0.12 & \underline{0.9231}    & 0.94 & 0.10 & \underline{0.9167}    & 0.93 & 0.11 & \underline{0.9084}    \\
0.07 & 0.92 & 0.08 & 0.9200          & 0.90 & 0.08 & 0.9091          & 0.86 & 0.06 & 0.8958          & 0.87 & 0.08 & 0.8952          \\
0.08 & 0.86 & 0.04 & 0.9053          & 0.87 & 0.06 & 0.9016          & 0.83 & 0.05 & 0.883           & 0.84 & 0.07 & 0.8806          \\
0.09 & 0.74 & 0.04 & 0.8315          & 0.74 & 0.05 & 0.8268          & 0.72 & 0.04 & 0.8205          & 0.74 & 0.05 & 0.8310       \\
\bottomrule
\end{tabular}}\label{table:A4}
\end{table*}

\section{Influence of Attack Method on the Determination of the Threshold}\label{appendix:threshold}
\SLJ{To study how the attack method affects the determination of the threshold, we report the TPR, FPR, and F1 scores using adversarial examples generated by different attack methods under different thresholds in \ref{table:threshold}. 
We show the best F1 scores in bold, second-best in italic, and third-best in underline, from which we can see that TextDefense performs the best when the threshold is in the range of 0.04 to 0.06.
Thus, the local attack method will not affect the determination of the threshold. 
In other words, we can directly pick a value between 0.04 and 0.06 as the threshold of TextDefense. 
}

\section{Influence of the Number of Local Samples on the Determination of the threshold}\label{appendix:threshold2}
\SLJ{We also experiment on the influence of the number of local samples on TextDefense's effectiveness, as shown in Table \ref{table:A4}. 
Similarly, we show the best F1 scores in bold, second-best in italic, and third-best in underline, from which we can see that TextDefense performs the best when the threshold is in the range of 0.04 to 0.06.
That is, the number of local clean samples and their corresponding adversarial examples will not affect the determination of the threshold. 
}

\begin{figure*}[t]
    \centering
    \subfloat[\SLJ{Amazon}]{
             \includegraphics[scale=0.4]{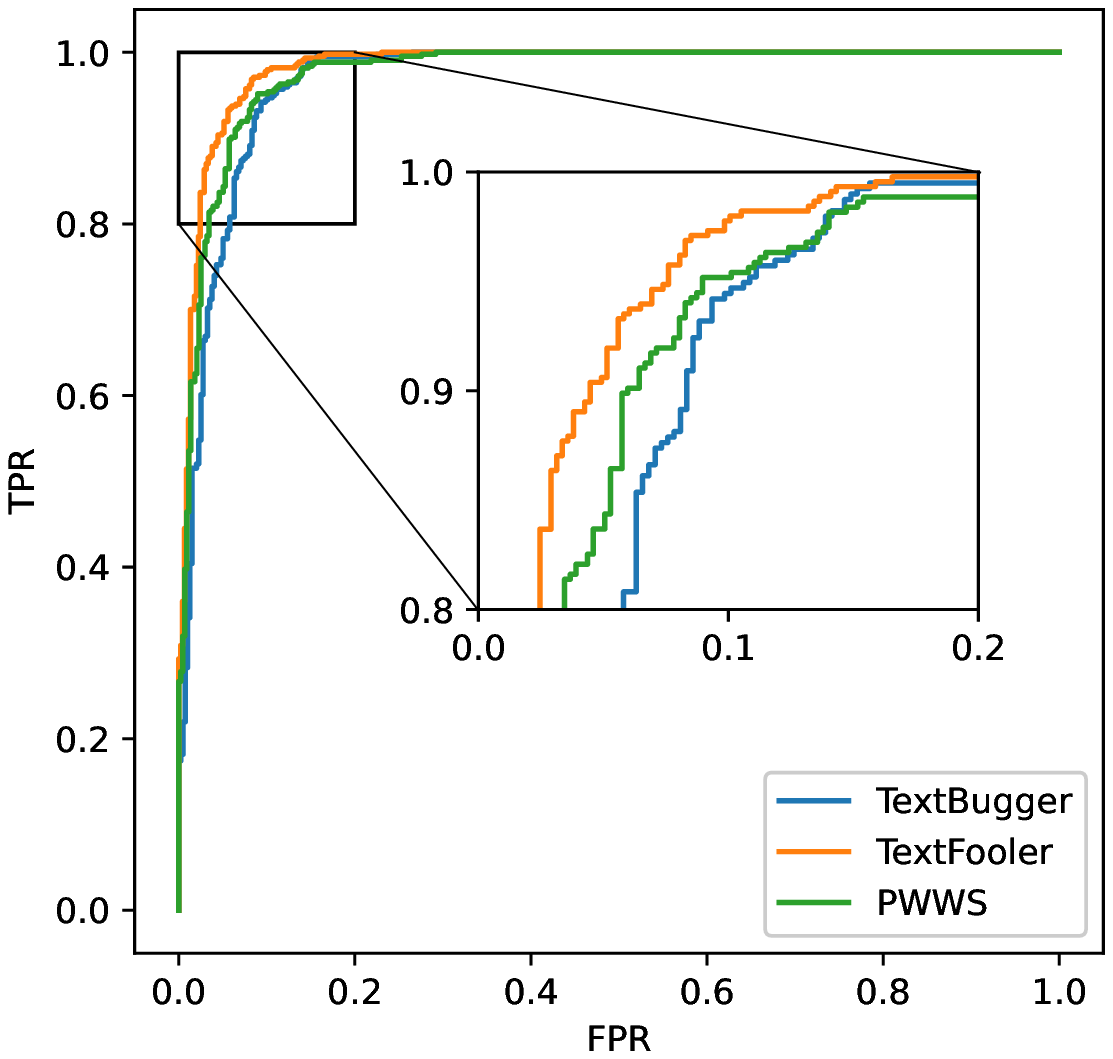}\label{subfig:to_amazon}}
    \subfloat[\SLJ{Yelp}]{
             \includegraphics[scale=0.4]{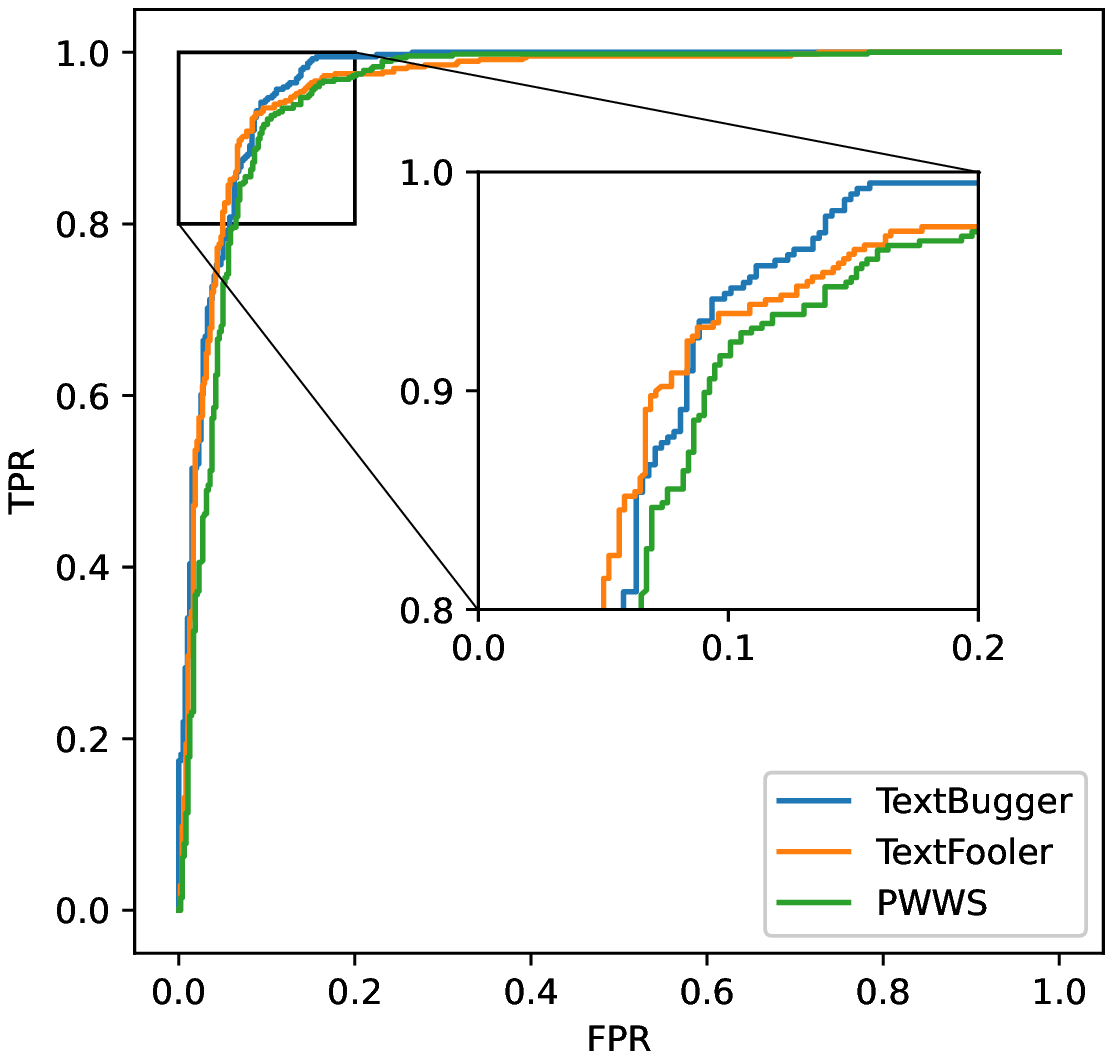}\label{subfig:to_yelp}}
    \subfloat[\SLJ{IMDb}]{
             \includegraphics[scale=0.4]{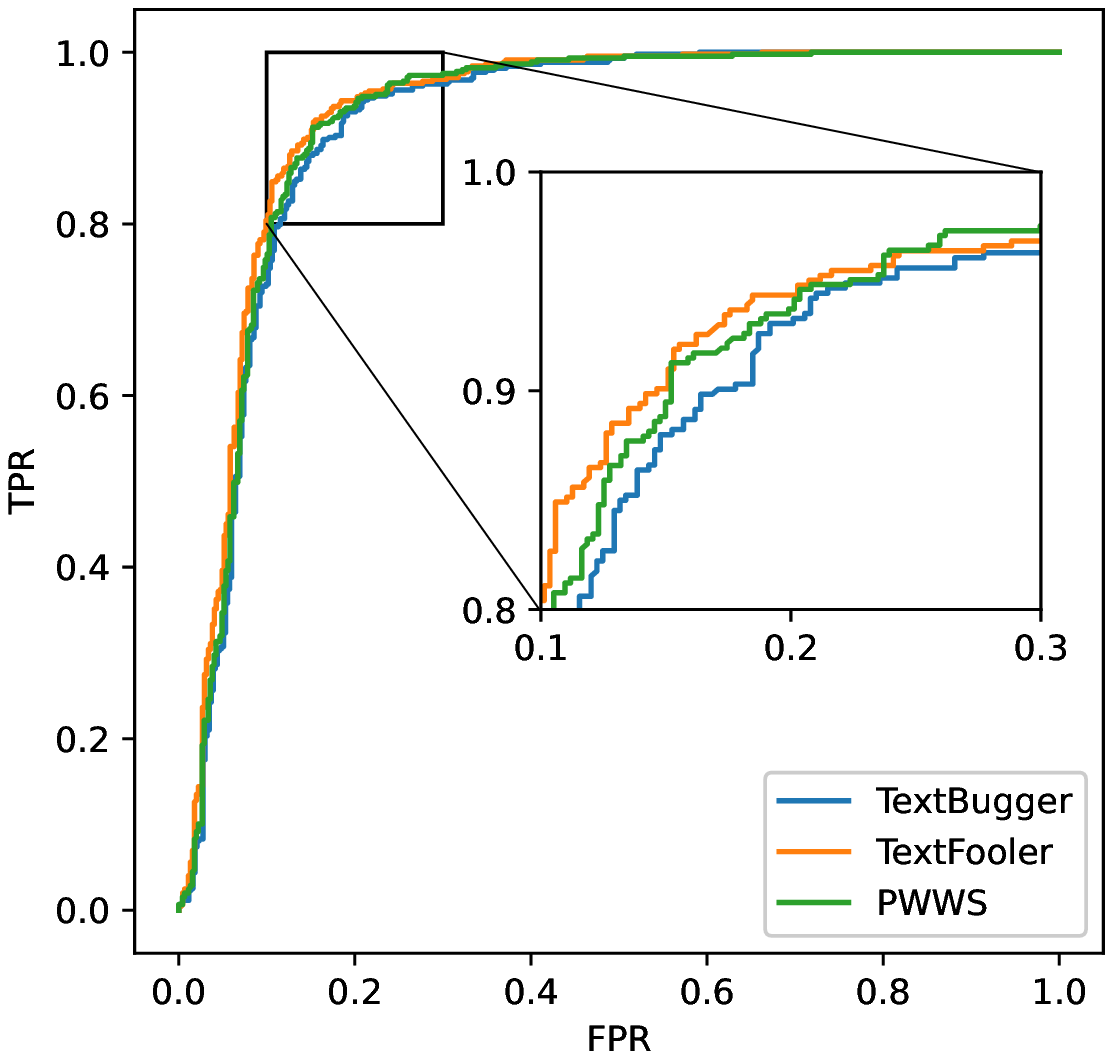}\label{subfig:to_imdb}}\\
    \subfloat[\SLJ{Twitter}]{
             \includegraphics[scale=0.4]{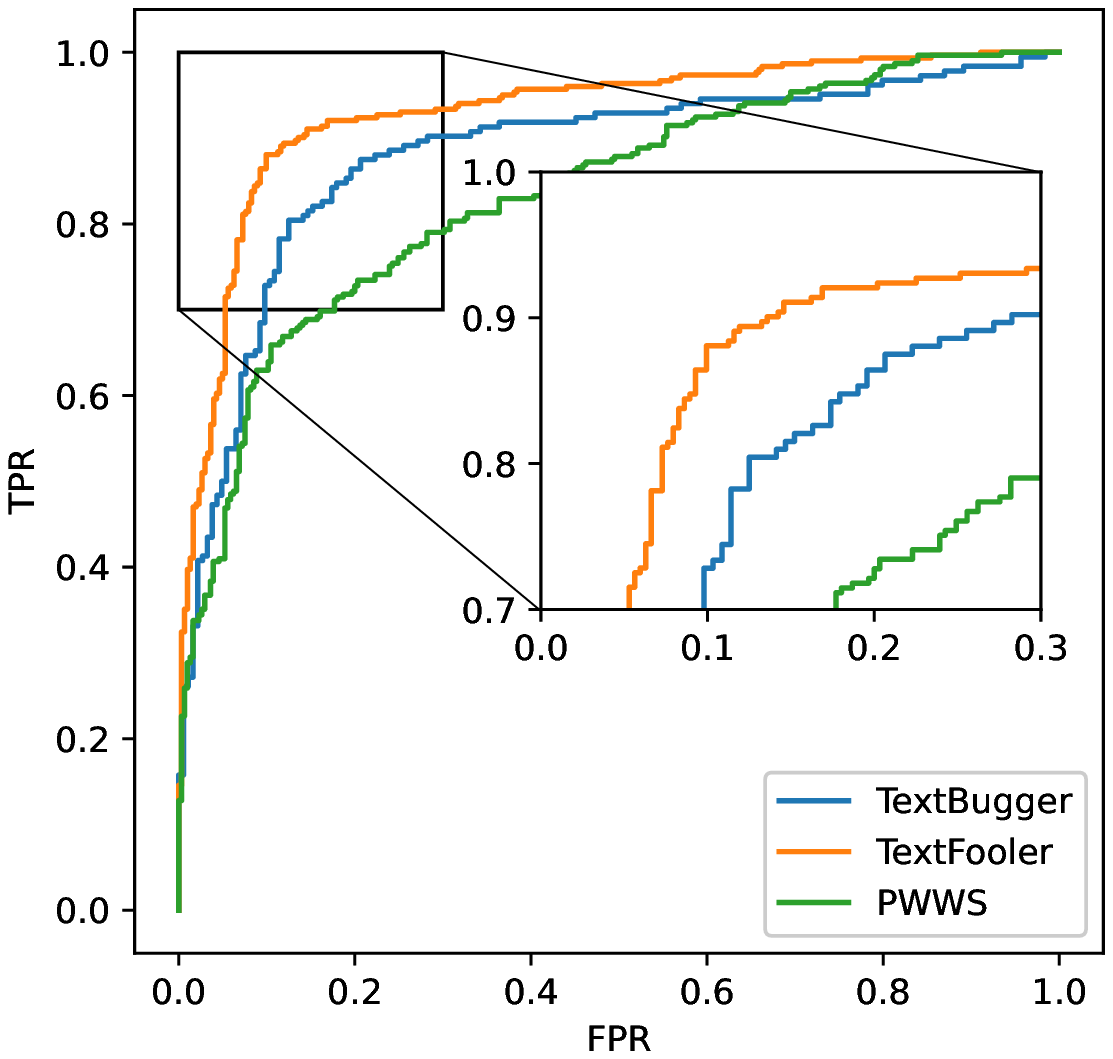}\label{subfig:to_twitter}}
    \subfloat[\SLJ{Jigsaw}]{
             \includegraphics[scale=0.4]{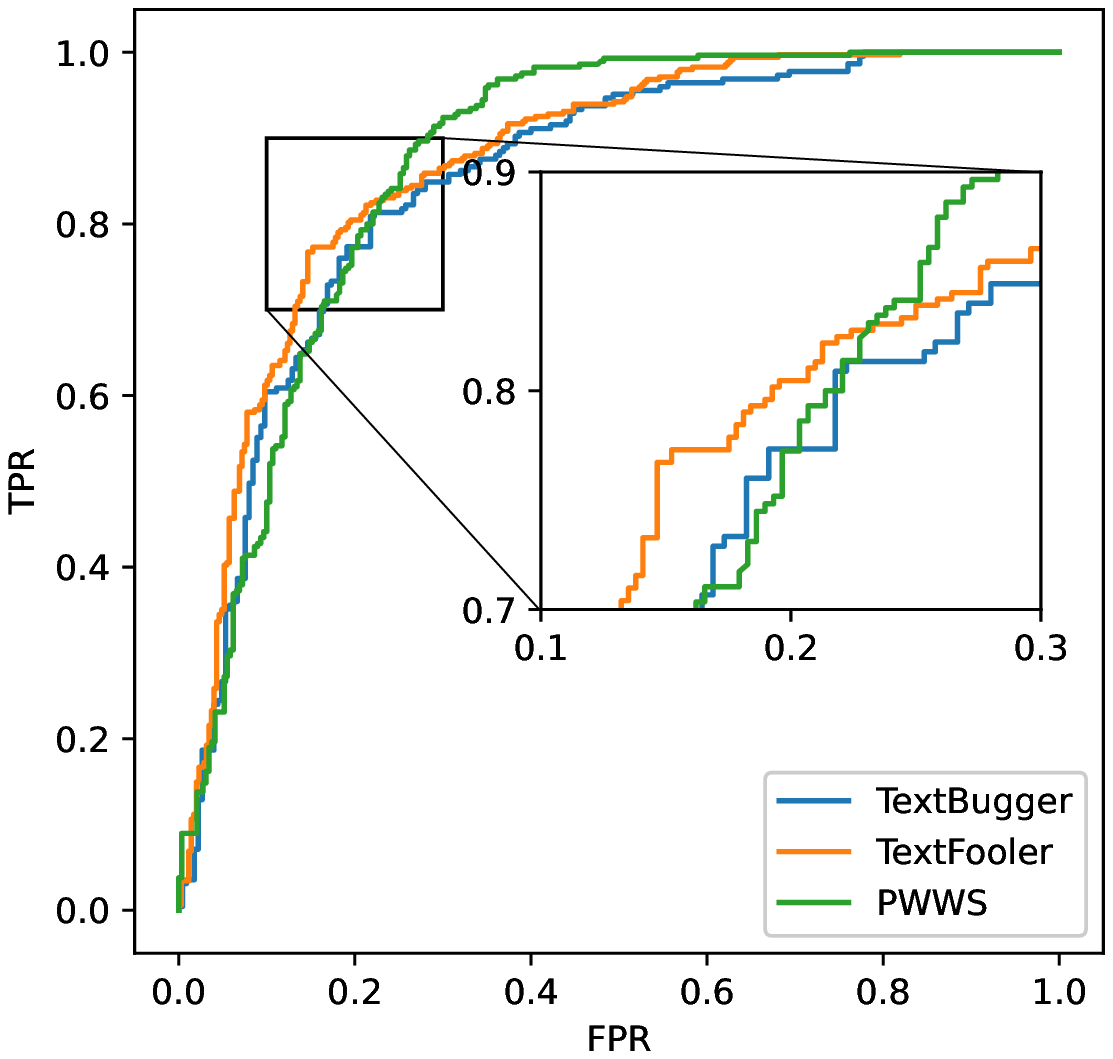}\label{subfig:to_jigsaw}}
         \caption{\SLJ{Trade-off between TPR and FPR of five datasets.}}
         \label{fig:tradeoff}   
\end{figure*}

\section{Trade-off between TPR and FPR}\label{appendix:trade-off}
\SLJ{
We have provided the trade-off between TPR and FPR (i.e., ROC curve) of the five datasets in Fig. \ref{fig:tradeoff}. 
We can see from Fig. \ref{subfig:to_amazon},\ref{subfig:to_yelp} and \ref{subfig:to_imdb} that TextDefense can achieve high TPR with low FPR. 
For example, in Fig. \ref{subfig:to_amazon}, it is appropriate to set a threshold that can achieve an FPR close to 0.1. 
At this time, TextDefense can detect 95\% of all adversarial examples. 
As shown in Table \ref{table:performance}, to achieve an FPR close to 0.1, we can set the threshold to 0.06.
For another example, in Fig. \ref{subfig:to_imdb}, it is appropriate to set a threshold that can achieve an FPR close to 0.15.
At this moment, TextDefense can detect 90\% of all adversarial examples. 
To achieve this FPR, we can set the threshold to 0.06.}

\SLJ{As for the Twitter and Jigsaw datasets, it is appropriate to set the threshold to 0.07 for them.
At this time, TextDefense can achieve an average TPR of 0.83 and an average FPR of 0.22 in the Jigsaw dataset.
For the Twitter dataset, with a threshold of 0.07, TextDefense can achieve a TPR of 0.86 and an FPR of 0.2 in the TextBugger attack, a TPR of 0.91 and an FPR of 0.16 in the TextFooler attack, and a TPR of 0.68 and an FPR of 0.14 in PWWS attack.
Since PWWS generates low-quality adversarial examples, TextDefense does not perform as expected in PWWS.}

\section{Performance of TextDefense and TextShield on Other Adversarial Attack Methods}\label{appendix:other_attack}
We also implement other adversarial attack methods in NLP to generate adversarial texts. 
To be specific, we use BERT-Attack \cite{li2020bert}, BAE \cite{garg2020bae}, DeepWordBug \cite{gao2018black}, IGA \cite{pmlr-v161-wang21a} and Kuleshov et al. \cite{kuleshov2018adversarial}.
Other works which either attack with low ASR or generate texts with low quality are not considered. 
For TextShield, we use the T5 model used in Sec. \ref{sec:comparison} to convert adversarial inputs.
We report the AUC of TextDefense and TextShield in Table \ref{tab:other_attack}

\begin{table}[h]
    \centering
    \caption{The performance of TextDefense and TextShield on other attack methods.}
    \SLJ{
    \scalebox{0.75}{\begin{tabular}{cccccc}
    \toprule
    Attck            & ASR    & LA & Similarity & TextDefense & TextShield    \\
                \hline
    BERT-Attack  & 97.64\%  & 0.8180 & 0.9616 & 0.8476 & 0.6414\\
    BAE         & 40.68\% & 0.7000 & 0.9663 & 0.9271 & 0.6522\\
    DeepWordBug  & 79.87\% & 0.3005 & 0.8521 & 0.9628 & 0.9238\\
    IGA         & 75.42\% & 0.3680 & 0.8787 & 0.9734 & 0.7226\\
    Kuleshov et al.   & 96.52\% & 0.5885 & 0.9232 & 0.9874 & 0.9299\\
    \bottomrule
    \end{tabular}}}
    \label{tab:other_attack}
\end{table}

From Table \ref{tab:other_attack}, we can see that TextDefense achieves high AUC on different attack method. 
Meanwhile, TextShield can only detect adversarial examples generated from DeepWordBug and Kuleshov et al. 
However, TextShield fails to perform well in BERT-Attack, BAE and IGA.
This indicates that the performance of TextShield highly relies on the diversity of the training samples of the NMT model.
When a new type of attack is proposed, TextShield would fail to detect the adversarial examples.

\begin{table}[h]
\centering
\caption{\SLJ{The average run-time per sample and extra storage used by different defense method.}}
\scalebox{0.9}{
\begin{tabular}{cccccc}
\toprule
              & DISP   & FGWS   & TextShield & TextDefense \\
              \hline
Run-time (s)     & 0.0317 & 3.9061 & 1.888      & 0.2227          \\
Storage (MB) & 876    & 78349  & 242        & 0          \\
\bottomrule         
\end{tabular}}\label{tab:runtime}
\end{table}

\begin{table}[h]
\centering
\caption{\SLJ{The average run-time per sample using different sampling numbers.}}
\SLJ{
\scalebox{0.77}{\begin{tabular}{ccccccc}
\toprule
sample         & 2      & 5      & 10     & 20     & 50     & ALL    \\
\hline
Amazon (87.54) & 0.0074 & 0.0146 & 0.0272 & 0.0518 & 0.118  & 0.2227 \\
Yelp (149.46)  & 0.0073 & 0.0147 & 0.0271 & 0.0517 & 0.1197 & 0.3487\\
\bottomrule
\end{tabular}}}\label{tab:runtime2}
\end{table}

\begin{table*}[t]
\begin{center}
\caption{The adversarial examples generated on Twitter and Jigsaw dataset.}
\vspace{-3mm}
	\scalebox{0.9}{\begin{tabular}{c c l r} 
		\toprule
        Dataset & Attack & \multicolumn{1}{c}{Text} & \multicolumn{1}{c}{Prediction}   \\
        \hline
        \multirow{2}{*}{Twitter} &\multirow{1}{*}{TextBugger}&  You're going by yourself because you have a problem you \bluetext{\st{fucking}} \redtext{fuc\(\kappa\)ing} \bluetext{\st{loser}} \redtext{lose}.  & 97\% pos $\rightarrow$ 98\% neg\\
        \cline{2-4}
        &\multirow{1}{*}{TextFooler}& don't care if it's real or fake just know that \bluetext{\st{ass}} \redtext{behinds} \bluetext{\st{fat}} \redtext{lard}. & 95\% pos $\rightarrow$ 98\% neg \\
        \hline
        \multirow{2}{*}{Jigsaw} &\multirow{1}{*}{TextBugger}& If you fidn this page then    == GO TO THELL \bluetext{\st{BITCH}} \redtext{BITH}! ==  & 98\% pos $\rightarrow$ 98\% neg\\
        \cline{2-4}
        &\multirow{1}{*}{TextFooler}& I am sorry, that was my \bluetext{\st{stupid}} \redtext{farcical} sister. it won't happen again! & 75\% pos $\rightarrow$ 97\% neg \\
		\bottomrule
	\end{tabular}}
\label{table:case1}
\end{center}
\end{table*}

\begin{table*}[t]
\begin{center}
\caption{The adversarial examples generated using TextFooler and PWWS on Amazon dataset.}
\vspace{-3mm}
	\scalebox{0.9}{\begin{tabular}{c l r} 
		\toprule
        Attack & \multicolumn{1}{c}{Text} & \multicolumn{1}{c}{Prediction}   \\
        \hline
        \multirow{3}{*}{TextFooler} &\makecell[l]{The Cambridge Companion to \bluetext{\st{Chaucer}} \redtext{Rabelais}: \bluetext{\st{Great}} \redtext{Whopping} \bluetext{\st{book}} \redtext{handout} that I am \\\bluetext{\st{using}} \redtext{exploiting} now. It is \bluetext{\st{very}} \redtext{terribly} \bluetext{\st{beneficial}} \redtext{preferable} to literature classes.} & 99\% pos $\rightarrow$ 81\% neg\\
        \cline{2-3}
        &\makecell[l]{\bluetext{\st{Great}} \redtext{Large} \bluetext{\st{product}} \redtext{products}!: \bluetext{\st{These}} \redtext{Those} salt and \bluetext{\st{pepper}} \redtext{onion} grinders are \bluetext{\st{great}} \redtext{large}. \\\bluetext{\st{Stylish}} \redtext{Overdressed} and \bluetext{\st{practical}} \redtext{veritable}, they both \bluetext{\st{work}} \redtext{acted} \bluetext{\st{perfectly}} \redtext{rather}.}& 100\% pos $\rightarrow$ 84\% neg \\
		\hline
		\multirow{2}{*}{PWWS} & \makecell[l]{\bluetext{\st{Happy}} \redtext{Felicitous} Neil fan!: I would consider this Neil \bluetext{\st{record}} \redtext{book} \bluetext{\st{one}} \redtext{unitary} of his most \\ successful and artistically satisfying \bluetext{\st{works}} \redtext{sour}.} & 99\% pos $\rightarrow$ 87\% neg\\
        \cline{2-3}
		&\makecell[l]{\bluetext{\st{Great}} \redtext{Bully} product!: These salt and pepper grinders are \bluetext{\st{great}} \redtext{capital}. Stylish and practical, \\ they both work \bluetext{\st{perfectly}} \redtext{dead}.} & 100\% pos $\rightarrow$ 82\% neg\\
		\bottomrule
	\end{tabular}}
\label{table:case2}
\end{center}
\end{table*}

\section{Run-time and Extra Storage Comparison}\label{appendix:runtime}
\SLJ{We calculate the run-time per query and the extra storage consumption for each method, and the result is shown in Table \ref{tab:runtime}. The table shows that FGWS takes 3.9s per sample and requires 78GB of extra storage to store the similarity matrix. TextShield takes an average of 1.888s per sample to generate the corrected text. TextShield, in our reimplementation, depends on the T5 model, which takes 242MB of storage. DISP takes 0.0317s per sample and utilizes two BERT models, which in total require 876MB storage. TextDefense takes 0.2227s per sample and requires no extra storage. }

\SLJ{In Appendix \ref{sample}, we have also experimented with the sampling technique, which shows high defense effectiveness and low computation cost. 
Thus, we also study the run-time of TextDefense using different sampling numbers. 
The run-time of TextDefense on Amazon and Yelp datasets with different sampling numbers is reported in Table \ref{tab:runtime2}. 
The average text length of Amazon and Yelp are 87.54 and 149.46, respectively.
The result shows that TextDefense can achieve 27.2ms per sample when sampling 10 words in a text, which exceeds DISP's speed. 
In addition, TextDefense takes only 7.4ms per sample when sampling two words from the text, which is much faster than other methods.
}

\section{Worst-case Analysis}\label{case_study}
In previous studies, we have found that the performance of TextDefense has slightly degraded on the Twitter and Jigsaw datasets, whose task is toxic comment detection. 
In the meantime, we have discovered that the two synonym substitution attacks (i.e., TextFooler and PWWS) have higher ASR with less perturbed words and can bypass TextDefense using higher constraints. 
In the following, we elaborate on the analysis of these two cases. 

\subsection{Twitter/Jigsaw}\label{case_twitter}
In Sec. \ref{sec:clean_performance}, we have speculated that the performance degradation on Twitter and Jigsaw datasets is the intrinsic property of the toxic detection task where abusive words can be easily detected. 
Therefore, we come up with the question that 
\textit{does adversarial attack on Twitter and Jigsaw replace abusive words with neutral words to achieve the attack?}
We illustrate four typical adversarial examples and their original examples in Table \ref{table:case1} where the perturbed words are shown in blue in the original text and red in the adversarial text. 

We can see that their original texts contain several abusive words in both datasets. 
However, in the adversarially generated examples, these abusive words are replaced either by character-level attack or `synonym' substitution.
In the abusive detection task, only if the abusive words are replaced by some unknown token like the generated `fuc$\kappa$ing' using TextBugger or replaced by some neutral synonyms like the `farcical' in TextFooler, does the model's prediction change. 
However, the replacement of the neutral synonym, though it maintains the syntax correctness, is not suitable in attacking the text from Twitter and Jigsaw due to the alteration of the sentence meaning.
This lead to the poor performance of TextDefense on abusive detection, where the generated texts are intrinsically not abusive anymore.
Nevertheless, as for datasets like Twitter and Jigsaw, deploying a spelling checker to defend against character-level attacks is crucial.

\subsection{TextFooler/PWWS}\label{case_tf}
Previous experiments have shown that TextFooler and PWWS can always attack the model with higher ASR and fewer perturbation words. 
Then, \textit{can we assume that the TextFooler and PWWS are stronger than TextBugger?}
Unfortunately, as we analyze the generated adversarial examples using TextFooler and PWWS, we find that most of them change the text's original meaning, which violates the definition of adversarial example.
We illustrate the adversarial examples and their original examples in Table \ref{table:case2}. 

In the first text, all positive emotional words `great' and `very beneficial' are converted into `whopping' and `terribly preferable', respectively, which significantly changes the original meaning of the text. 
Similarly, in the second text, all positive emotional words `great', `Stylish', and `perfectly' are converted into `large', `Overdressed', and `rather', respectively. 
Since the words `whopping', `large', and `rather' are emotionally neutral, while the words `terribly' and `overdressed' are emotionally negative, the resulting texts are predicted to be negative. 
Such generated examples no longer satisfy the definition of adversarial examples. 
We have observed similar results in PWWS where emotionally positive words `great' and `perfectly' in the second text are converted into emotionally negative words `bully' and `dead'. 
However, there are no existing methods to quantify changes in sentence meaning. 
Therefore, current adversarial attack methods modify the semantically meaningful words arbitrarily, ignoring the definition of adversarial example. 

Thus, we consider TextDefense is still effective in defending `real' adversarial examples.

\section{The Distribution of Perturbations Identified by DISP}\label{appendix:disp}
\begin{figure}[h]
    \centering
    \includegraphics[scale=0.15]{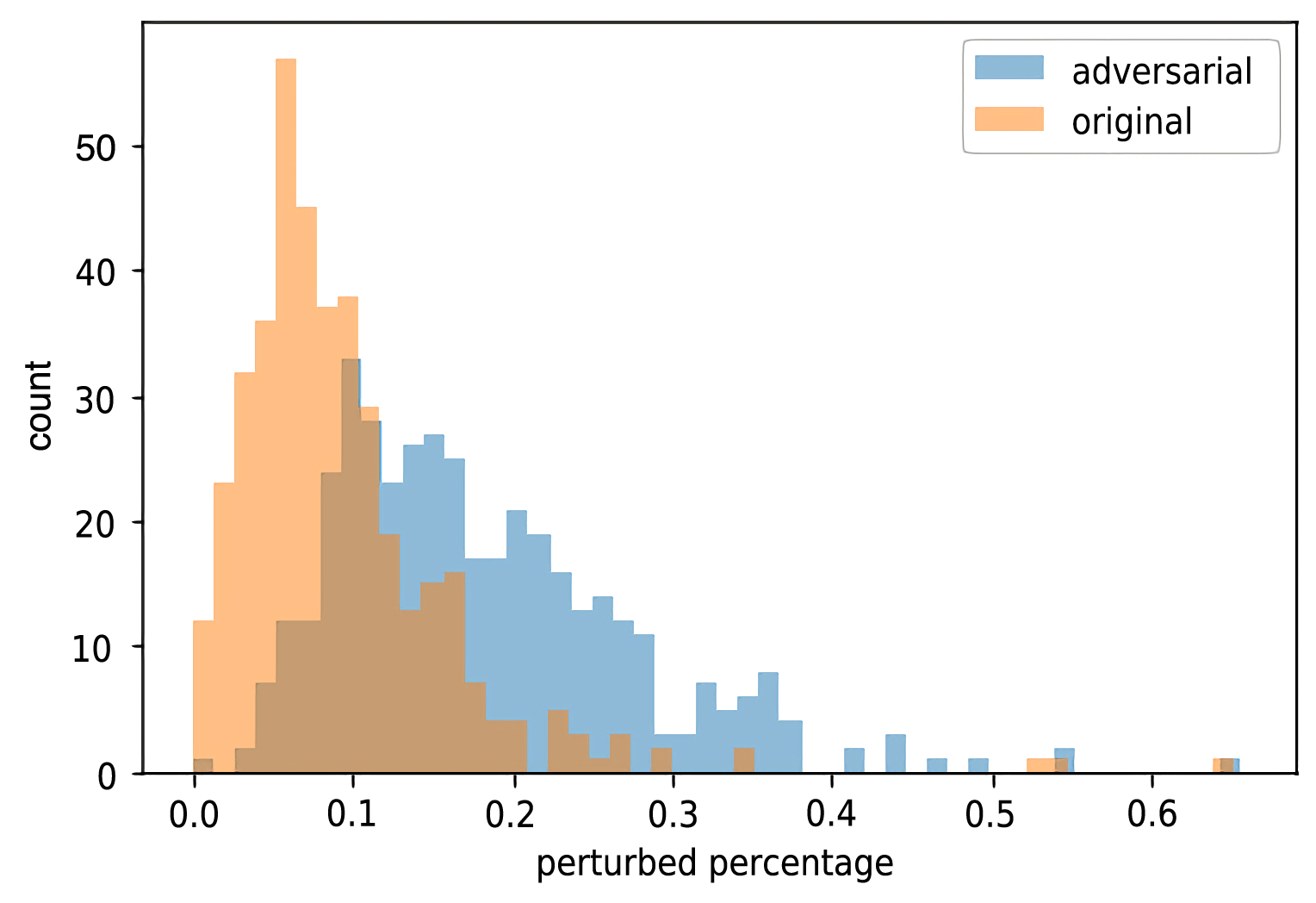}
    \caption{The distributions of the percentage of perturbed words from adversarial examples and original examples using DISP.}
    \label{fig:disp}
\end{figure}

In Fig. \ref{fig:disp}, we show the distribution of perturbations identified by DISP on clean and adversarial examples..
We can see that the distribution of adversarial examples is more scattered and the values are larger comparing to the distribution of clean examples. However, there is a huge overlap between the two distributions. 
Therefore, using the percentage of the perturbed words can hardly distinguish the adversarial examples.

\section{The Adversarial Attack Rules under PWWS}\label{appendix:pwws_rule}
\SLJ{Several word replacements in false negatives of adversarial Jigsaw and IMDb texts under PWWS attack are shown in Table \ref{tab:jigsaw_pwws} and \ref{tab:imdb_pwws}. 
Table \ref{tab:jigsaw_pwws} shows that toxic words are all replaced by non-toxic words in the PWWS attack. 
In addition, words like `make', `know', `mean', etc., in non-toxic texts are replaced by toxic words that failed to be detected by TextDefense.
Table \ref{tab:imdb_pwws} shows that sentiment-negative words like `bad', `worst', `awful' are replaced by emotion-neutral words like `tough', `awesome', `spoilt'.
Similarly, sentiment-positive words like `loved', `best', `good' are replaced by emotion-neutral words like `screw', `full', `dependable'.
These attacks result in a higher false negative rate of TextDefense.
}

\begin{table}[t]
\caption{\SLJ{Several word replacement in false negatives of adversarial Jigsaw texts under PWWS attack.}}
\SLJ{
\begin{tabular}{crcl}
\toprule
\multicolumn{1}{l}{}                & original &  & adversarial             \\
\hline
\multirow{9}{*}{\makecell{toxic\\ to\\ non-toxic}} & stupid   & \rightarrowfill & stunned, dazed, stupefied \\
                                    & shit     & \rightarrowfill & stool, shop, hoot         \\
                                    & suck     & \rightarrowfill & draw, lactate            \\
                                    & fucking  & \rightarrowfill & eff, know, jazz      \\
                                    & ass      & \rightarrowfill & bottom, seat, fundament   \\
                                    & gay      & \rightarrowfill & jovial, festal, brave     \\
                                    & sucks    & \rightarrowfill & draw, absorb             \\
                                    & crap     & \rightarrowfill & make, stool, ca-ca        \\
                                    & fuck     & \rightarrowfill & eff, know, jazz           \\
                                    \hline
\multirow{5}{*}{\makecell{non-toxic\\ to\\ toxic}} & made     & \rightarrowfill & shit                    \\
                                    & know     & \rightarrowfill & fuck                    \\
                                    & make     & \rightarrowfill & shit                    \\
                                    & mean     & \rightarrowfill & bastardly               \\
                                    & can      & \rightarrowfill & crapper   				 \\
                                  \bottomrule             
\end{tabular}}\label{tab:jigsaw_pwws}
\end{table}

\begin{table}[]
\caption{\SLJ{Several word replacement in false negatives of adversarial IMDb text under PWWS attack.}}
\SLJ{
\begin{tabular}{crcl}
\toprule
                                      & original &                                & adversarial                   \\
                                      \hline
\multirow{4}{*}{\makecell{negative\\ to\\ positive}} & bad      & \rightarrowfill & tough, forged, spoilt         \\
                                      & worst    & \rightarrowfill & tough                         \\
                                      & awful    & \rightarrowfill & awesome, dreaded              \\
                                      & crap     & \rightarrowfill & make, ca-ca                   \\
                                      \hline
\multirow{4}{*}{\makecell{positive\\ to\\ negative}} & funny    & \rightarrowfill & laughable, fishy, rummy       \\
                                      & best     & \rightarrowfill & full, dependable, respectable \\
                                      & loved    & \rightarrowfill & screw                         \\
                                      & good     & \rightarrowfill & dependable, unspoilt, just   \\
                                      \bottomrule
\end{tabular}}\label{tab:imdb_pwws}
\end{table}

\end{document}